\documentclass[a4paper,12pt]{article}
\usepackage{amsmath, amssymb}
\usepackage{graphicx}
\usepackage{hyperref}
\usepackage{xcolor}
\usepackage{graphicx}
\usepackage{amsmath,amsfonts,amssymb,mathrsfs}
\usepackage{caption}
\usepackage{booktabs}
\usepackage{subcaption}
\usepackage{hyperref}
\usepackage[utf8]{inputenc}
\usepackage{algorithm}
\usepackage[noend]{algpseudocode}
\usepackage[left=2.3cm,right=2.3cm,top=2cm,bottom=2cm]{geometry}
\usepackage{tabularx} % à ajouter dans le préambule
\usepackage{graphicx}
\usepackage{subcaption}

\usepackage[style=numeric-comp,sorting=none,sortcites=true,giveninits=true,uniquename=false]{biblatex}
\newcommand{\bbGamma}{{\mathpalette\makebbGamma\relax}}
\newcommand{\makebbGamma}[2]{%
  \raisebox{\depth}{\scalebox{1}[-1]{$\mathsurround=0pt#1\mathbb{L}$}}%
}

\title{Predicting Grain Growth in Polycrystalline Materials Using Deep Learning Time Series Models}
\author{
    Eliane Younes, Elie Hachem, Marc Bernacki \\ 
    \\
    Mines Paris, PSL University\\
Centre for material forming (CEMEF), UMR CNRS\\ 
06904 Sophia Antipolis, France
}
\date{}

\begin{document}
\maketitle

\begin{abstract}
Grain Growth strongly influences the mechanical behavior of materials, making its prediction a key objective in microstructural engineering. In this study, several deep learning approaches were evaluated, including recurrent neural networks (RNN), long short-term memory (LSTM), temporal convolutional networks (TCN), and transformers, to forecast grain size distributions during grain growth. Unlike full-field simulations, which are computationally demanding, the present work relies on mean-field statistical descriptors extracted from high-fidelity simulations. A dataset of 120 grain growth sequences was processed into normalized grain size distributions as a function of time. The models were trained to predict future distributions from a short temporal history using a recursive forecasting strategy. Among the tested models, the LSTM network achieved the highest accuracy (above 90\%) and the most stable performance, maintaining physically consistent predictions over extended horizons while reducing computation time from about 20 minutes per sequence to only a few seconds, whereas the other architectures tended to diverge when forecasting further in time. These results highlight the potential of low-dimensional descriptors and LSTM-based forecasting for efficient and accurate microstructure prediction, with direct implications for digital twin development and process optimization.\
\end{abstract}

\noindent\textbf{Keywords:} Grain Growth, Deep Learning, Time Series Forecasting, Microstructural Evolution, Mean-Field Descriptors

\section{Introduction}

Although the grain growth (GG) mechanism has been known for more than one century and actively studied for 70 years~\cite{smith1948,burke1952,mullins1956,neumann1952}, it remains a very active research topic in terms of in-situ data acquisition~\cite{bhattacharya2021}, involved equations, and modelling \cite{rollett2017}. This interest can be directly related to the importance of predicting grain size during annealing of metallic materials, which acts as a first-order parameter in discussing certain mechanical properties such as flow stress or creep resistance~\cite{ZHANG2022161710,lasalmonie1986influence}. The historical background has brought us to the current global picture~\cite{herring1999}: GG in single-phase polycrystals relatively free of point and line defects is driven by the minimisation of grain boundary (GB) areas to decrease the system’s interfacial energy. The velocity $\mathbf{v}$
at which GBs migrate when subjected to a driving pressure $P$ is assumed to be the product of the GB mobility $\mu$ and $P\mathbf{n}$ with $\mathbf{n}$ the outward unitary normal to the GB. Following the chain rules clearly depicted by Herring in~\cite{herring1999}, the capillarity driving pressure linked to an evolving GB interface can be described, at the first order of the small displacement performed, as:

\begin{equation}\label{eq:herring1}
P \approx -\underbrace{\left(\gamma \pmb{\mathbb{Id}} + \nabla_{\mathbf{n}}\nabla_{\mathbf{n}}\gamma\right)}_{\pmb{\bbGamma}\left(\mathbf{n}\right)}:\pmb{\mathbb{K}},
\end{equation}
with $\pmb{\mathbb{K}}=\nabla \mathbf{n}$ the curvature tensor and $\pmb{\bbGamma}\left(\mathbf{n}\right)$ the GB stiffness tensor. Beyond the temperature dependence of the mobility, $\mu$ and $\gamma$, depend also on the underlying crystallography at GBs which are today depicted in the 5-dimensional parameter space of the misorientation of the abutting crystals and the GB plane (inclination). This leads to significant spatial heterogeneity of GB kinetic properties in polycrystalline materials~\cite{smith1948,kohara1958}. Moreover, generally, $\pmb{\bbGamma}\left(\mathbf{n}\right)$ and $\pmb{\mathbb{K}}$ are not diagonalizable in the same basis of orthonormal eigenvectors and the calculation of the tensor double product contracted is not straightforward and must be carefully evaluated in the right basis by also taking into account the dependence of $\gamma$ to crystallographic orientation symmetries~\cite{abdeljawad2018,du2007}.\\

Finally, when $\gamma$ is assumed not dependent of the inclination $\mathbf{n}$ (non-anisotopic), i.e. only dependent of the GB disorientation angle (heterogeneous) or homogeneous, Eq.\ref{eq:herring1} simplifies in the well-known mesoscopic kinetic equation for curvature flow:
\begin{equation}\label{eq:herring2}
P \approx -\gamma\left(\kappa_1 + \kappa_2\right)=-\gamma\kappa,\text{ and }\mathbf{v}\approx -\mu\gamma\kappa\mathbf{n},
\end{equation}

with $\left\{\kappa_1,\kappa_2\right\}$ the principal curvatures of the curvature tensor $\pmb{\mathbb{K}}$ and $\kappa=\kappa_1 + \kappa_2$ its trace, generally known as the mean curvature of an interface in materials science and, in a mathematically more consistent way, as twice the mean curvature in other communities such as fluid mechanics. In this formulation, the contributions of $\mu$ and $\gamma$ can no longer be clearly separated, and the term \textit{reduced mobility} has therefore become widely adopted in the literature to denote their product.\medbreak

Thanks to the improvement of GG experiments, this global picture is more and more discussed in the state of the art. Indeed, monitored in-situ in a synchrotron by 3D X-ray diffraction microscopy techniques, which avoid bias inherent to 2D observations, opens the way to precise reverse engineering in order to discuss the $\left\| \mathbf{v} \right\|/\kappa$ ratio ~\cite{chen2020,bhattacharya2021,xu2023,florez2022,lyu2025,yang2026,naghibzadeh2024,bizana2023,zhang2020}, i.e., the apparent reduced mobility of each GB while knowing their characteristics in the 5-dimensional parameter space of the misorientation and inclination. If Eq.\ref{eq:herring2} is indeed a simplification of lower scale phenomena in constant discussions, the interpretation of the results obtained tends in some cases to refute or at least to question the real impact of the curvature on the kinetics of the interfaces~\cite{bhattacharya2021} or to dispel the 5-parameter description of the reduced mobility~\cite{zhang2020}. The observation of interfaces not moving towards their center of curvature ($\left\| \mathbf{v} \right\|/\kappa$ ratio of opposite sign to the expected one) and/or presenting kinetics very different from other grain boundaries with similar characteristics in the 5-dimensional space indicate that further investigation is required.
However, in view of the different proposed equations, the sometimes used recent strong statements that "GG driving pressure is not correlated to curvature" is inherently ambiguous. Are we speaking of the trace of the curvature tensor (Eq.\ref{eq:herring2}) or the curvature tensor itself (Eq.\ref{eq:herring1})? In the state of the art, it is ultimately Eq.\ref{eq:herring2} that is often questioned without taken into account the inclination dependence in experimental or numerical analysis. It is, in fact, both expected and reassuring that Eq.\ref{eq:herring2} does not hold at the local scale. However, this should not be interpreted as evidence against its validity in simulations involving a large number of interfaces. A major pitfall in recent literature dedicated to GG is probably the growing tendency to consider Eq.\ref{eq:herring2} as obsolete, not only for describing individual boundaries but even for modeling polycrystalline grain growth in realistic annealing conditions and large systems. This is probably the main symptom emerging from recent literature on the subject, and it likely represents a misconception.\medbreak

Why, in a single-phase system—without second-phase particles and without solute drag at grain boundaries—do the classical models of Burke$\&$Turnbull~\cite{burke1952} and Hillert~\cite{zhang2020} for predicting the mean grain size remain so remarkably accurate in annealing predictions, even though they are fundamentally based on Eq.\ref{eq:herring2}? The answer is probably that, at the scale of a few hundred interfaces, Eq.\ref{eq:herring2} for kinetics-based models, combined with a heterogeneous description of $\gamma$ and an Arrhenius-type temperature dependence for $\mu$, remains extremely powerful. The same argument could be made for the limiting grain size predicted by the Smith–Zener pinning model~\cite{zener1949}. Finally, it seems important not to dismiss the valuable along with the questionable. Advancing toward Eq.\ref{eq:herring1} within anisotropic formalisms, not overlooking pressures often considered negligible that may locally become significant during the evolution towards an equilibrium, refining 4D experimental data, and of course improving current models at the GB scale, are topics of prime importance. However, this should not lead to the conclusion that Eq.\ref{eq:herring2} is therefore fundamentally not usable. For polycrystals that are not strongly textured and not dominated by coherent twins, Eq.\ref{eq:herring2} continues to provide an exceptionally robust and predictive framework for modeling grain growth at the mesoscale through kinetics frameworks — its simplicity being, paradoxically, the key to its enduring success. So, one can totally agree with the conclusion of Yang of coworkers \cite{yang2026} where it is claimed that Eq.\ref{eq:herring2} should definitively not be applied locally to understand the behavior of individual GBs in 4D experimental data or anisotropic full-field simulations but it remains that this equation can be used without hesitation for kinetics-based models in large simulation systems.\medbreak

That said about the physics of the problem, high-fidelity modeling at the polycrystalline scale represents an important strategy for simulating GG and has been employed for more than four decades~\cite{himanen2019data}. Several stochastic and deterministic approaches have been developed for GG modeling, including the Monte Carlo Potts (MCP) method~\cite{wu1982potts, peczak1993monte}, phase-field models~\cite{chen1997computer, krill2002computer}, cellular automata~\cite{liu1996simulation, hesselbarth1991simulation}, front-tracking techniques~\cite{frost1988two, barrales2007novel}, and level-set methods~\cite{elsey2009diffusion, zhao1996variational, bernacki2008level}. Among front-tracking methods \cite{bernacki2024vertex}, the ToRealMotion (TRM) approach is a Lagrangian framework that efficiently captures GG kinetics; its performance has been validated on large experimental datasets, including heterogeneous and anisotropic systems \cite{florez2020novel,florez2022}. This explains our choice concerning this recent numerical framework to generate the dataset used in this work.\\

Thus, full-field simulations have proven to be highly valuable tools for investigating microstructural evolution. Despite their predictive capabilities, extending these models to industrial or macroscopic scales remains challenging, as it would require extremely high computational resources and simulation times that increase rapidly with the system size. This limitation has motivated the development of simplified yet physically grounded modeling strategies capable of bridging the micro- and macro-scales at a lower computational cost.\\

Among these strategies, physically based metallurgical mean-field models offer an effective compromise between accuracy and numerical efficiency \cite{Gourdet2003,Montheillet2009,perez2008,Maire2018,seret2020,Cram2009,Bernard2011}. This simplification drastically reduces computational requirements and makes mean-field simulations well suited for process design or large-scale studies. Nevertheless, as highlighted in the work of Roth et al. \cite{roth2023comparison}, mean-field approaches, although considerably faster than full-field models, can still involve significant computational effort when applied to realistic industrial cases, owing to the need to solve numerous coupled equations for large populations of representative grains. Therefore, although mean-field models represent a major step toward computationally tractable simulations, their use in real-time or large-scale applications remains limited, calling for new strategies capable of further accelerating microstructural predictions.\\

To overcome these computational limitations, recent research has turned toward data-driven methods such as machine learning (ML) and, more specifically, deep learning (DL)~\cite{shi2022time, sonata2024comparison}.
ML algorithms can learn complex nonlinear relationships directly from data, enabling fast and accurate predictions without explicitly solving the underlying physical equations. When trained on mean-field data generated from full-field or mean-field simulations, these models can reproduce microstructural evolution with much lower computational cost while maintaining physical consistency. 
This makes DL promising complementary tools for accelerating microstructure prediction and, ultimately, for enabling efficient digital twin applications.\\

Building on these advances, several recent studies have applied ML to accelerate the prediction of microstructural evolution using full-field simulation data. Ahmad et al.~\cite{ahmad2023accelerating} proposed a hybrid autoencoder–ConvLSTM model designed to forecast successive microstructural frames from phase-field simulations. The architecture combines convolutional layers to extract spatial features with recurrent layers to capture temporal dependencies, allowing accurate reconstruction of grain morphology evolution. As a result, they achieved substantial reductions in computational time comparatively to phase-field simulations while maintaining fine spatial resolution and morphological accuracy. Qin et al.~\cite{qin2023grainnn} developed \textit{GrainNN}, a sequence-to-sequence LSTM framework enhanced with attention mechanisms to reproduce grain evolution under various physical conditions. The model learns temporal relationships directly from full-field datasets and efficiently captures dynamic grain growth. Their model demonstrated superior performance in predicting grain size distribution and morphology compared to conventional simulation techniques, with significant gains in computational efficiency. More recently, Tep et al.~\cite{TEP2025121486} introduced a ConvLSTM–autoencoder model specifically tailored for grain growth prediction from large-scale full-field datasets. Their proposed method resulted in a speedup of computation times by a factor close to 90 comparatively to classical simulations while preserving the fidelity of grain morphology, boundary topology, and size distribution through a dedicated loss function. Despite these advances, the reliance on high-dimensional image-based data remains a limitation, as generating and processing such datasets demand extensive computational resources and long training times. To overcome these constraints, recent studies have started to explore mean-field representations as an alternative source of training data, offering a balance between physical consistency and computational efficiency.\\

Jha et al.~\cite{jha2017combined} combined CALPHAD thermodynamic simulations with a k-Nearest Neighbors (k-NN) regression model to predict nanocrystal size and volume fraction in soft magnetic alloys, using only global processing parameters such as temperature and annealing time. Their approach achieved accurate predictions of key microstructural features while drastically reducing the required computational time. Similarly, Khandelwal et al.~\cite{khandelwal2021surrogate} used homogenized descriptors—including dislocation density, phase fraction, and local stress—to train an LSTM model that captures history-dependent deformation in dual-phase microstructures. The resulting model successfully reproduced the material’s stress–strain behavior under cyclic loading, demonstrating the potential of ML-assisted mean-field frameworks for history-dependent problems. Fu et al.~\cite{fu2022prediction} further developed an ensemble machine learning model trained on mean-field simulation outputs to predict grain size and recrystallization fraction under varying thermo-mechanical processing conditions. Their method achieved robust and accurate predictions across a wide range of parameters, confirming the effectiveness of combining ML with mean-field simulations for property prediction.\\

% Despite demonstrating the potential of ML trained on mean-field data, these studies primarily focused on predicting static, scalar properties, such as final grain size, phase fraction, or recrystallization degree, rather than capturing the continuous evolution of microstructures. In contrast, the present work aims to extend this concept to the dynamic modeling of grain growth. Specifically, we propose an ML-based framework capable of forecasting the temporal evolution of grain size distributions derived from validated full-field simulations. By treating these distributions as time-dependent sequences, the approach enables the application of deep sequence modeling architectures to capture the kinetics of grain growth. This provides an intermediate route between traditional mean-field formulations and computationally demanding full-field models, offering a scalable, physically consistent, and efficient framework for dynamic microstructure forecasting and digital twin development.\\

Despite demonstrating the potential of machine learning trained on mean-field data, these studies primarily focus on predicting static, scalar properties. In contrast, our work shifts attention to the dynamic modeling of microstructural evolution, aiming to capture the temporal behavior of grain size distributions derived from validated full-field simulations. By treating these distributions as evolving sequences, we enable the application of sequence modeling techniques to represent complex, time-dependent material behaviors. This approach provides an intermediate solution between traditional scalar-based predictions and the high-resolution, computationally expensive full-field methods, offering a scalable and efficient alternative for dynamic microstructure forecasting.\\

Based on this idea, the objective of the present work is to forecast grain size distribution using DL models designed for time series data. The dataset employed originates from TRM simulations, previously validated against experimental measurements, with each simulation represented as a sequential time series. Several architectures suited for sequence forecasting are examined, including Recurrent Neural Networks (RNN), Long Short-Term Memory networks (LSTM)\cite{elsworth2020time}, Temporal Convolutional Networks (TCN)\cite{hewage2020temporal}, and Transformer models based on attention mechanisms~\cite{vaswani2017attention}. The paper is organized as follows: first, the data preparation and forecasting framework based on mean-field descriptors is introduced. Then, the DL models are presented and compared in terms of predictive accuracy and efficiency, followed by a description of the experimental setup and evaluation procedure. Finally, the results are analyzed and discussed with regard to model performance and potential for real-time applications.

\section{Methodology}
In this study, GG is treated as a temporal modeling problem, with the objective of learning and predicting the evolution of grain size distributions over time. The methodology is organized into three main steps. First, a physically consistent dataset of grain growth sequences is generated using validated simulation frameworks. Second, the raw outputs are processed into compact time-series representations suitable for machine learning. Finally, a forecasting strategy is developed, in which historical microstructural states are used to predict subsequent grain evolution. This pipeline provides a data-driven framework for analyzing microstructural dynamics during thermal treatments.

\subsection{Dataset Description}
The dataset used in this study is composed of sequential textual data describing the main grain growth descriptors (grain size distribution) recorded at different temporal points during microstructural evolution.
The data generation process followed a structured sequence: definition of simulation parameters, generation of initial microstructures, and simulation of their temporal evolution. It began by specifying key parameters for constructing initial configurations using a Laguerre–Voronoï Tessellation algorithm (LaVoGen) \cite{hitti2012precise}.
Four domain sizes were considered (2$\times$2, 3$\times$3, 4$\times$4, and 5$\times$5 mm$^2$), which influenced both the number of grains and their interactions. Grain size distributions were initialized using a normal distribution $\mathcal{N}$, with the mean fixed at 20 $\mu$m and the standard deviation varied from 2 to 32 $\mu$m in integer steps. By maintaining a constant mean and expressing grain sizes in normalized form $R/\langle R \rangle$, the dataset ensures both compactness and generalizability , enabling applicability across diverse material systems and grain size scales.\\

\begin{figure}[ht]
  \centering
  \includegraphics[width=1\linewidth]{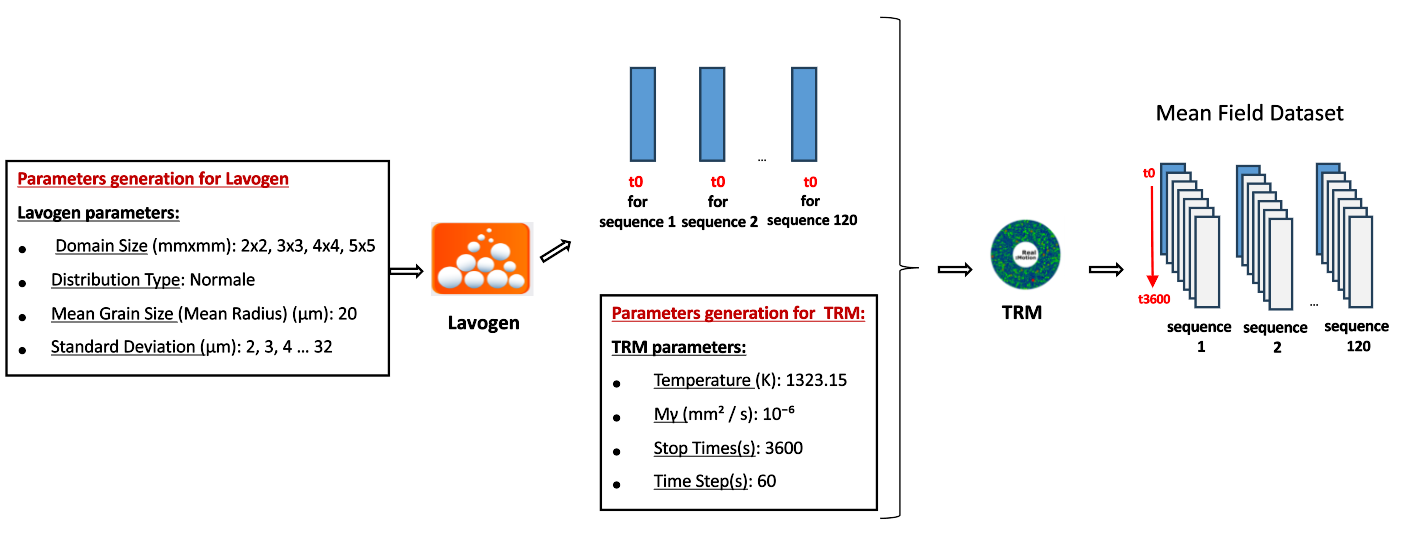}
  \caption{Pipeline for generating microstructure evolution sequences using  Laguerre-Voronoï tessellation (Lavogen) and TRM simulations.}
  \label{pipline1}
\end{figure}

Once the initial configurations were established, they were introduced into the \textit{TRM} simulation framework \cite{florez2020novel}, which employs a front-tracking method to model grain boundary migration during annealing. All simulations were performed at a fixed temperature of 1323.15 K with a maximal reduced mobility of $10^{-6}$ mm$^2$/s, a Read-Shockley description of $\gamma$ \cite{read1950} and a random cristallographic orientation for grains, i.e. with representative parameters of 304L stainless steel under this temperature \cite{maire2017}. These parameters were selected to reduce complexity while preserving the essential physics of grain growth. The simulation timelines can be rescaled to describe systems with different kinetics (temperature and mobility). For instance, a system with twice the mobility would reach the same microstructural state in half the time. This property extends the applicability of the dataset beyond the original conditions. The data generation workflow is summarized in Figure~\ref{pipline1}.\\

Each TRM-generated sequence includes both visual and statistics outputs. However, the focus of this work is on the grain size distribution. It provides a description of the microstructure at each recorded step, allowing the tracking of how the overall grain size distribution evolves. For every minute of simulated evolution, the dataset records the equivalent circular radius of each grain ($ECR$) which corresponds to the radius of the circle of same area that the grain. This structured representation supports both grain-level tracking and statistical analysis of microstructural dynamics over time. Moreover, it preserves the essential physical features of grain growth and remains computationally efficient, making it well-suited for machine learning applications.\\

The complete dataset consists of 120 evolution sequences, each describing the transformation of a microstructure during one hour of annealing, with simulation outputs recorded every minute. Generating the full dataset required approximately 40 hours of computation, with each sequence requiring around 20 minutes.

\subsection{Data Preprocessing}
To prepare the raw data generated by TRM for input into the neural network, it was necessary to extract and organize the most relevant features that would support the learning process. The primary focus was on the evolution of grain size distributions. To reduce dimensionality and storage requirements, grain sizes were grouped into discrete intervals (bins). At each time step, the number of grains within each interval was counted to construct a frequency distribution.\\

The frequencies were then normalized by dividing each bin count by the total number of grains, resulting in a probability distribution:

\begin{equation}
\hat{f}_i = \frac{f_i}{\sum_{j=1}^{n} f_j}
\tag{2}
\end{equation}

This normalization ensured that the data remained consistent across time steps and allowed the model to focus on relative distributions rather than absolute quantities. As a result, each time step was represented by a fixed-length vector capturing the probabilistic grain size distribution.\\

The number of intervals was varied between 10 and 50 to evaluate the effect of resolution. Different configurations were tested to identify a bin count that preserved sufficient structural detail without introducing instability. The resulting vectors formed the time-series input to the neural network, enabling it to learn the dynamics of microstructural evolution.\\

The dataset was then divided into three subsets: training, validation, and testing. Several data-splitting ratios were examined, including 80:15:5, 80:10:10, 70:20:10, and 70:15:15 (training:validation:test). These specific ratios were chosen to analyze the influence of the amount of training data and validation size on forecasting accuracy. The training set was used to optimize the model parameters, the validation set served to tune hyperparameters and monitor generalization, and the test set provided an unbiased evaluation of predictive performance on unseen data. Configurations with a larger training portion (e.g., 80\%) were intended to ensure sufficient learning of the underlying microstructural dynamics, while those with larger validation and test portions (e.g., 20\% or 15\%) were included to evaluate the model’s robustness against overfitting.\\

At a first step, the forecasting task was designed to simulate a one-hour thermal treatment. For each simulation, the model received the first five minutes of grain size data and was required to predict the subsequent microstructural evolution over the remaining duration. This setup evaluated the model’s capability to forecast future grain growth behavior from limited initial information. Cross-validation with the different data-splitting strategies ensured robust assessment of stability and performance.\\

This preprocessing step was essential to reduce noise, improve computational efficiency, and ensure that the neural network received structured and meaningful input for accurate training and prediction of grain growth dynamics.

\subsection{Time Series Forecasting Strategy}

\begin{figure}[ht]
  \centering
  \includegraphics[width=0.9\linewidth]{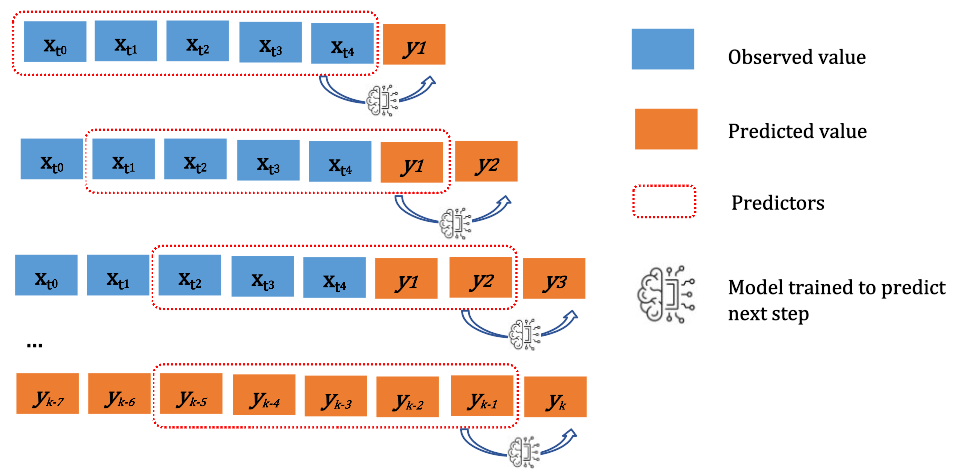}
  \caption{Recursive Forecasting Using a Fixed-Length Sliding Window}
  \label{sliding_window2}
\end{figure}

Using historical data, DL models can learn how past values in a time series relate to future values \cite{lara2020temporal}. In our approach, forecasting is performed using a fixed-length input window of five time steps (each corresponding to one minute), ${t_0, t_1, t_2, t_3, t_4}$. The model predicts the value at the following step, $t_5$, denoted as $y_1$. To maintain a consistent input size, a sliding window strategy is applied, in which the window shifts forward after each prediction (Fig.~\ref{sliding_window2}) \cite{shen2020novel,qin2023grainnn}.\\

The first prediction can be expressed as:
\begin{equation}
y_1 = f(x_{t_0}, x_{t_1}, x_{t_2}, x_{t_3}, x_{t_4})
\tag{3}
\end{equation}

where $f(\cdot)$ is the nonlinear mapping learned by the network, linking five consecutive inputs to one output.

Forecasting continues recursively, with each new prediction replacing the oldest input. The second prediction $y_2$ corresponds to time $t_6$ and is obtained from ${x_{t_1}, x_{t_2}, x_{t_3}, x_{t_4}, y_1}$:

\begin{equation}
y_2 = f(x_{t_1}, x_{t_2}, x_{t_3}, x_{t_4}, y_1)
\tag{4}
\end{equation}
Similarly, the third prediction $y_3$ corresponds to $t_7$, using ${x_{t_2}, x_{t_3}, x_{t_4}, y_1, y_2}$:

\begin{equation}
y_3 = f(x_{t_2}, x_{t_3}, x_{t_4}, y_1, y_2),\quad \text{and so on.}
\tag{5}
\end{equation}

In general, after several iterations the model becomes fully autoregressive, relying exclusively on its previous outputs:

\begin{equation}
y_k = f(y_{k-5}, y_{k-4}, y_{k-3}, y_{k-2}, y_{k-1})
\tag{6}
\end{equation}

This recursive strategy allows the model to autonomously generate sequences of future values beyond the initial input window, without requiring additional observed data.

\section{Deep Learning Frameworks}
To capture the temporal dynamics of grain growth, several DL architectures commonly used for time-series forecasting were investigated and compared. These models were selected for their ability to learn from sequential data and to represent different mechanisms of temporal dependency. An overview of each framework is provided below.

\subsection{Recurrent Neural Networks (RNN)}

\begin{figure}[ht]
  \centering
  \includegraphics[width=0.8\linewidth]{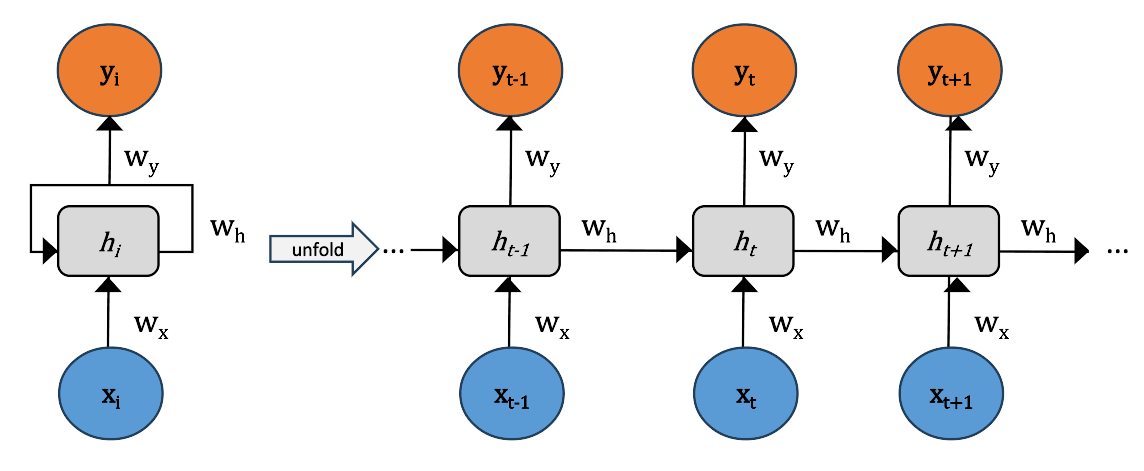}
  %\caption{Structure of the RNN \cite{shi2022time}}
  \caption{Structure of the RNN}
  \label{rnn}
\end{figure}

RNNs \cite{sherstinsky2020fundamentals} are widely used for modeling sequential data, as they retain information across time steps through a hidden state mechanism \cite{lecun2015deep}. At time step \( t \), the output depends on the current input \( x_t \) and the previous hidden state \( h_{t-1} \) \cite{wu2020deep}. As illustrated in Fig.~\ref{rnn}, the RNN updates its internal state sequentially, combining new inputs with historical context.\\

The RNN model is parameterized by three weight matrices \( W_x \), \( W_h \) and \( W_y \), in addition to two bias vectors \( b_s \) and \( b_y \), which together determine how the input and state information are transformed \cite{bengio1994learning}. The hidden state and output are computed as:

\[
\begin{aligned}
h_t &= \tanh(W_{xh} \cdot (x_t \oplus h_{t-1}) + b_s) \\
y_t &= \sigma(W_y \cdot h_t + b_y)
\end{aligned} \tag{7}
\]
\\

where \( x_t \in \mathbb{R}^m \) denotes the input vector containing \( m \) features at time step \( t \); 
\( W_{xh} \in \mathbb{R}^{n \times (m+n)} \) and \( W_y \in \mathbb{R}^{n \times n} \) are the learnable weight matrices, 
with \( n \) representing the number of neurons in the hidden layer. 
The bias vectors \( b_h, b_y \in \mathbb{R}^n \) correspond to the hidden state and output, respectively. 
The activation function \( \sigma \) is a sigmoid nonlinearity, and \( h_t \) denotes the hidden state. 
The operator \( \oplus \) indicates the concatenation of the current input \( x_t \) with the previous hidden state \( h_{t-1} \).
\\

While RNNs are effective for capturing short-term dependencies, they struggle with long sequences due to the vanishing gradient problem \cite{le2016quantifying, ribeiro2020beyond}, which limits their ability to retain information from earlier time steps.

\begin{figure}[ht]
  \centering
  \includegraphics[width=0.6\linewidth]{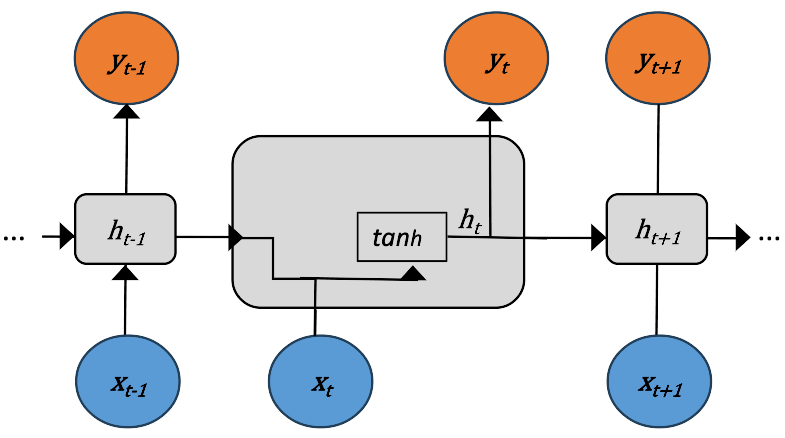}
  \caption{Computation process describing the RNN}
  \label{rnn2}
\end{figure}

\subsection{Long Short-Term Memory (LSTM)}
\begin{figure}[ht]
  \centering
  \includegraphics[width=0.7\linewidth]{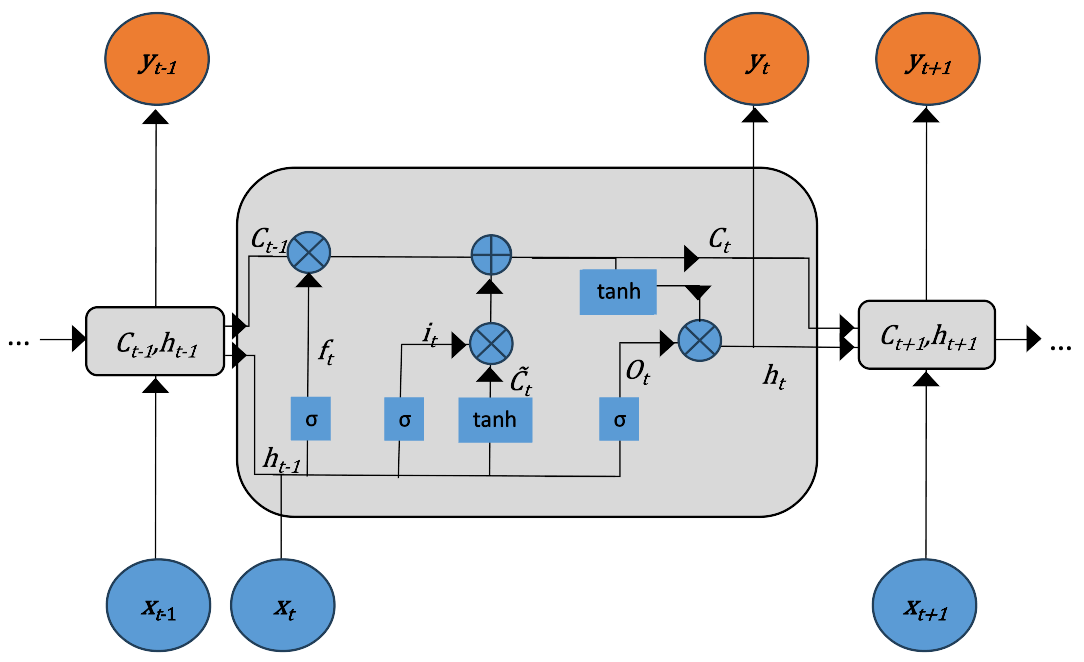}
  \caption{Computation process involved in an LSTM}
  \label{lstm}
\end{figure}

LSTM networks~\cite{yu2019review} extend RNNs by introducing gating mechanisms to mitigate the vanishing gradient issue and capture long-term dependencies \cite{hochreiter1997long}. Each unit maintains both a hidden state \( h_t \) and a cell state \( C_t \), which are updated based on the current input \( x_t \) and the previous states (Fig. \ref{lstm}).\\

The gates regulate information flow as follows:

\[
\begin{aligned}
f_t &= \sigma(W_f \cdot (x_t \oplus h_{t-1}) + b_f) \\
i_t &= \sigma(W_i \cdot (x_t \oplus h_{t-1}) + b_i) \\
\tilde{C}_t &= \tanh(W_C \cdot (x_t \oplus h_{t-1}) + b_C) \\
C_t &= f_t \cdot C_{t-1} + i_t \cdot \tilde{C}_t \\
o_t &= \sigma(W_o \cdot (x_t \oplus h_{t-1}) + b_o) \\
h_t &= \tanh(C_t) \cdot o_t \\
y_t &= \sigma(W_y \cdot h_t + b_y)
\end{aligned} \tag{8}
\]

Where \( W_f, W_i, W_C, W_o \in \mathbb{R}^{n \times (m+n)} \), and \( W_y \in \mathbb{R}^{n \times n} \) are weight matrices, and \( b_f, b_i, b_C, b_o, b_y \in \mathbb{R}^n \) are bias vectors.

By selectively forgetting, updating, and outputting information, LSTMs are able to retain relevant features over long sequences, making them particularly effective for time series forecasting \cite{saxena2021introduction, srivastava2017essentials}.

\subsection{Transformer Networks}

Transformer models~\cite{vaswani2017attention} employ attention mechanisms to model temporal dependencies without recurrence. An encoder-only design is adopted in this work, combining sinusoidal positional encodings with causal self-attention and feed-forward layers, following approaches such as the Time Series Transformer~\cite{zerveas2021transformer}.\\

\begin{figure}[ht]
\centering
\includegraphics[width=0.5\linewidth]{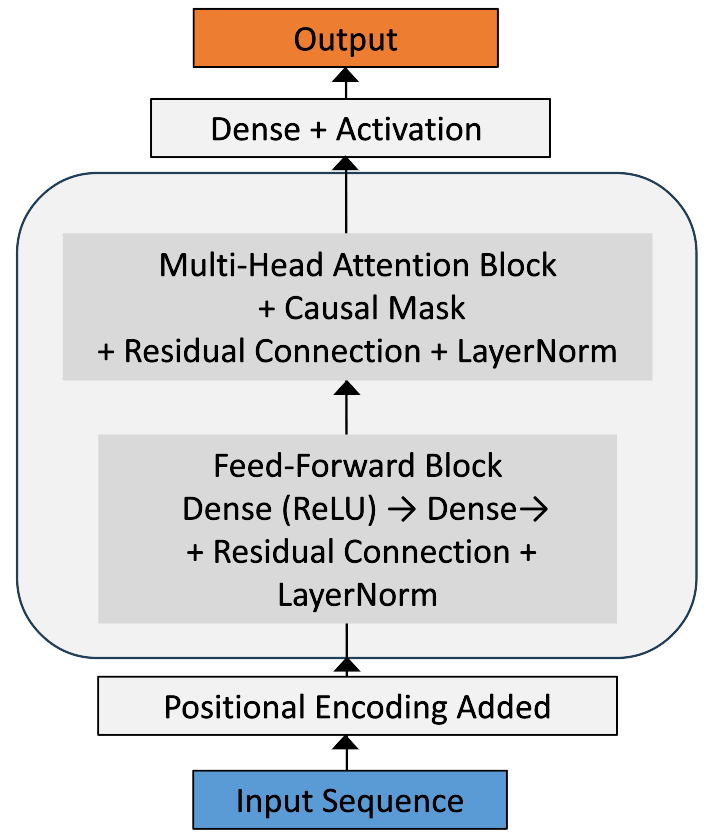}
\caption{Structure of the Transformer-based forecasting model}
\label{transformer}
\end{figure}

Ass shown in Fig.~\ref{transformer}, a fixed-length input sequence is first augmented with positional encodings to preserve temporal order. The sequence then passes through stacked layers of masked multi-head self-attention and feed-forward sub-blocks, each followed by residual connections and layer normalization. The causal mask ensures autoregressive forecasting by restricting attention to present and past steps. The final hidden state is used to predict the next value.\\

Transformers enable parallel computation and capture long-range dependencies efficiently \cite{alammar2018illustrated, harvard2018annotated}. However, when data are limited or dependencies are primarily local, they may not consistently outperform recurrent models such as LSTMs.

\subsection{Temporal Convolutional Network (TCN)}

Temporal Convolutional Networks (TCNs)~\cite{gopali2021comparison} apply 1D convolutions to sequential data, learning temporal patterns without recurrence. Unlike LSTMs, which use internal memory states, TCNs rely on causal and dilated convolutions to capture both short- and long-range dependencies \cite{bai2018empirical}.\\

\begin{figure}[ht]
\centering
\includegraphics[width=0.5\linewidth]{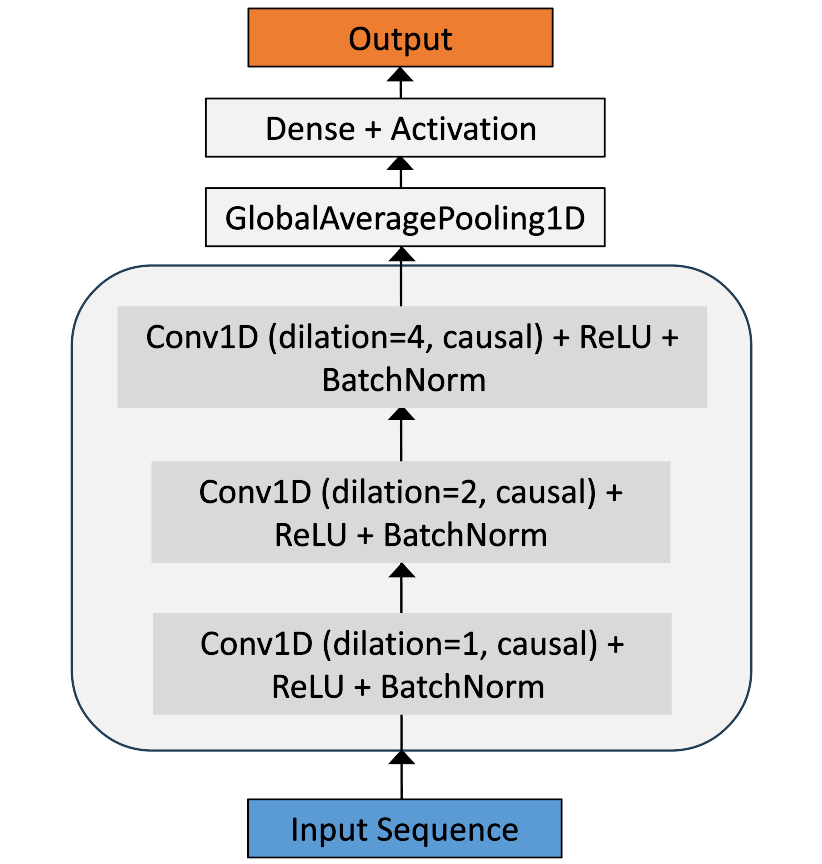}
\caption{Structure of the TCN-based forecasting model}
\label{tcn}
\end{figure}

In this work, the TCN architecture consists of three stacked 1D convolutional layers with dilation rates of 1, 2, and 4, respectively. Causal padding ensures that the output at time 
$t$ depends only on inputs up to $t$. Each layer is followed by ReLU activation and batch normalization. A \texttt{GlobalAveragePooling1D} layer summarizes the sequence into a fixed-size vector, which is passed through a dense layer with ReLU activation to produce the prediction (Fig.~\ref{tcn}).\\

Formally, a dilated causal convolution is defined as:

\begin{equation}
\text{Conv1D}(x)(t) = \sum_{i=0}^{k-1} w(i) \cdot x(t - d \cdot i)
\tag{9}
\end{equation}

where $w(i)$ are learnable filter weights, $k$ is the kernel size, and $d$ is the dilation factor that expands the receptive field exponentially across layers. This design allows the model to efficiently capture both local and distant dependencies in the input sequence.

\section{Experimental Setup and Training}

A series of controlled experiments were performed to evaluate the forecasting performance of the proposed DL models. This section describes the model configurations, training procedures, and evaluation metrics used in the study.

\subsection{Hyperparameters}

A preliminary hyperparameter search was performed to evaluate the effect of learning rate, batch size, and optimizer choice on model performance. Learning rates in the range $10^{-5}$–$10^{-3}$ and batch sizes of 32, 64, 128, 256, and 512 were tested using the Adam optimizer \cite{kingma2017adammethodstochasticoptimization}. Based on validation results, a configuration combining Adam with a learning rate of $10^{-4}$ and model-specific batch sizes was selected, as it provided a good balance between convergence speed, stability, and accuracy (see  Table~\ref{tab:dl_hyperparams}). Specifically, smaller batch sizes (32–128) were found to improve generalization and stability for complex architectures such as the Transformer and LSTM, while slightly larger batches offered faster convergence for simpler models like RNN. Smaller learning rates led to very slow convergence, while higher values caused training instability and oscillations, particularly for recurrent and transformer-based models. Each model was trained for 300 epochs, which ensured full convergence without signs of overfitting, as verified by the validation loss behavior. To ensure fair comparison, this configuration was applied uniformly in terms of learning rate and optimizer, with differences limited only to the batch size to account for model complexity. The adopted hyperparameters are summarized in Table~\ref{tab:dl_hyperparams}.\\

\begin{table}[h]
\centering
\caption{Training hyperparameters used for each deep learning model}
\label{tab:dl_hyperparams}
\begin{tabular}{c c c c c c}
\hline
\textbf{Model} & \textbf{Activation} & \textbf{Optimizer} & \textbf{LR} & \textbf{Epochs} & \textbf{Batch} \\
\hline
RNN &  &  &  & & 128 \\
LSTM & ReLU & Adam & $10^{-4}$ & 300 & 64 \\
TCN &  &  &  &  & 64 \\
Transformer &  & & &  & 32 \\
\hline
\end{tabular}
\end{table}

Each architecture was adapted to the sequential nature of the dataset. The RNN and LSTM models were implemented with two stacked recurrent layers of 128 hidden units, followed by a dense output layer. To limit overfitting, a dropout rate of 0.20 was applied to these two models only. For temporal convolutional modeling, the TCN architecture consisted of three dilated 1D convolutional layers with dilation factors of 1, 2, and 4, combined with Batch Normalization (BN), global average pooling, and a dense layer. BN was employed to stabilize and accelerate training by normalizing the activations across each mini-batch, which improved gradient flow and convergence. The Transformer model included two encoder blocks, each composed of multi-head self-attention and feed-forward layers, with positional encoding to preserve sequence order. In all cases, the output layer produced a 30-dimensional prediction vector, activated with ReLU. The detailed network structures are reported in Table~\ref{tab:dl_architectures}.

\begin{table}[h]
\centering
\caption{Network architectures used for each model}
\label{tab:dl_architectures}
\begin{tabular}{c p{12cm}}
\hline
\textbf{Model} & \textbf{Architecture} \\
\hline
RNN & SimpleRNN(128, return\_sequences=True) $\rightarrow$ SimpleRNN(128) $\rightarrow$ Dropout(0.20) $\rightarrow$ Dense(30) $\rightarrow$ ReLU \\
\hline
LSTM & LSTM(128, return\_sequences=True) $\rightarrow$ LSTM(128) $\rightarrow$ Dropout(0.20) $\rightarrow$ Dense(30) $\rightarrow$ LeakyReLU \\
\hline
TCN & Conv1D(64, kernel=3, dilation=1) $\rightarrow$ BN $\rightarrow$ Conv1D(64, dilation=2) $\rightarrow$ BN $\rightarrow$ Conv1D(64, dilation=4) $\rightarrow$ BN $\rightarrow$ GAP $\rightarrow$ Dense(30) $\rightarrow$ ReLU \\
\hline
Transformer & Input + PositionalEncoding $\rightarrow$ 3 $\times$ (MultiHeadAttention(20 heads, causal) $\rightarrow$ Add+Norm $\rightarrow$ FeedForward(128) $\rightarrow$ Add+Norm) $\rightarrow$ Dense(30) $\rightarrow$ LeakyReLU \\
\hline
\end{tabular}
\end{table}

This unified experimental framework allows performance differences to be attributed mainly to the architectural design of each model, ensuring a consistent and equitable basis for comparison.

\subsection{Evaluation Metrics}

Model performance was evaluated using three error metrics: Root Mean Square Error (RMSE), Mean Absolute Error (MAE), and Mean Relative Error (MRE). 
Unlike traditional point forecasting, where each time step is represented by a single scalar value \cite{reza2022multi}, our model outputs a full distribution of values across multiple bins at each time step. Accordingly, all metrics are computed by considering the differences between the predicted and actual distributions across all bins and time steps.\\

RMSE quantifies the standard deviation of prediction errors across all bins and time steps, reflecting the typical magnitude of deviations between predicted and true frequencies:

\begin{equation}
\text{RMSE} = 
\sqrt{
\frac{1}{T \times B} 
\sum_{t=1}^{T} \sum_{b=1}^{B} 
(\hat{y}_{t,b} - y_{t,b})^2
}
\tag{10}
\end{equation}

MAE measures the average absolute deviation across all bins, providing an interpretable measure of overall distributional accuracy:

\begin{equation}
\text{MAE} = 
\frac{1}{T \times B} 
\sum_{t=1}^{T} \sum_{b=1}^{B} 
|\hat{y}_{t,b} - y_{t,b}|
\tag{11}
\end{equation}

MRE expresses the error relative to the true value at each bin, averaged over all bins and time steps. This scale-invariant metric facilitates comparisons between distributions of different magnitudes:

\begin{equation}
\text{MRE (\%)} = 
\frac{100}{T \times B} 
\sum_{t=1}^{T} \sum_{b=1}^{B} 
\left| 
\frac{\hat{y}_{t,b} - y_{t,b}}{y_{t,b}} 
\right|
\tag{12}
\end{equation}

In these equations, \( T \) denotes the number of time steps, \( B \) the number of bins per distribution, \( y_{t,b} \) the observed frequency at bin \( b \) and time \( t \), and \( \hat{y}_{t,b} \) its corresponding predicted value.

\section{Results and Discussion}
\begin{figure}[htb]
    \centering

    % First row
    \begin{subfigure}[b]{0.45\textwidth}
        \centering
        \includegraphics[width=\textwidth]{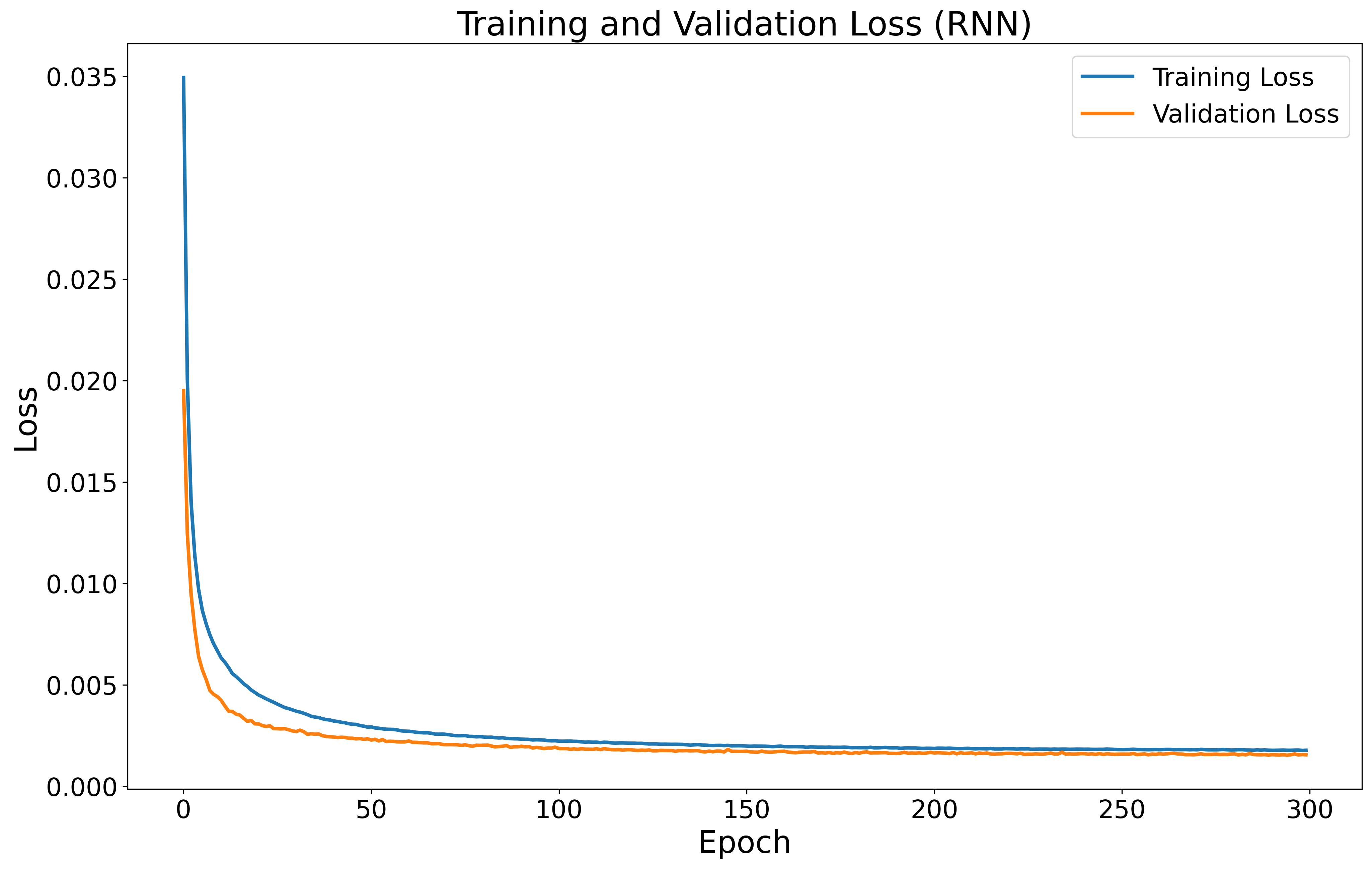}
        \caption{RNN}
    \end{subfigure}
    \hfill
    \begin{subfigure}[b]{0.45\textwidth}
        \centering
        \includegraphics[width=\textwidth]{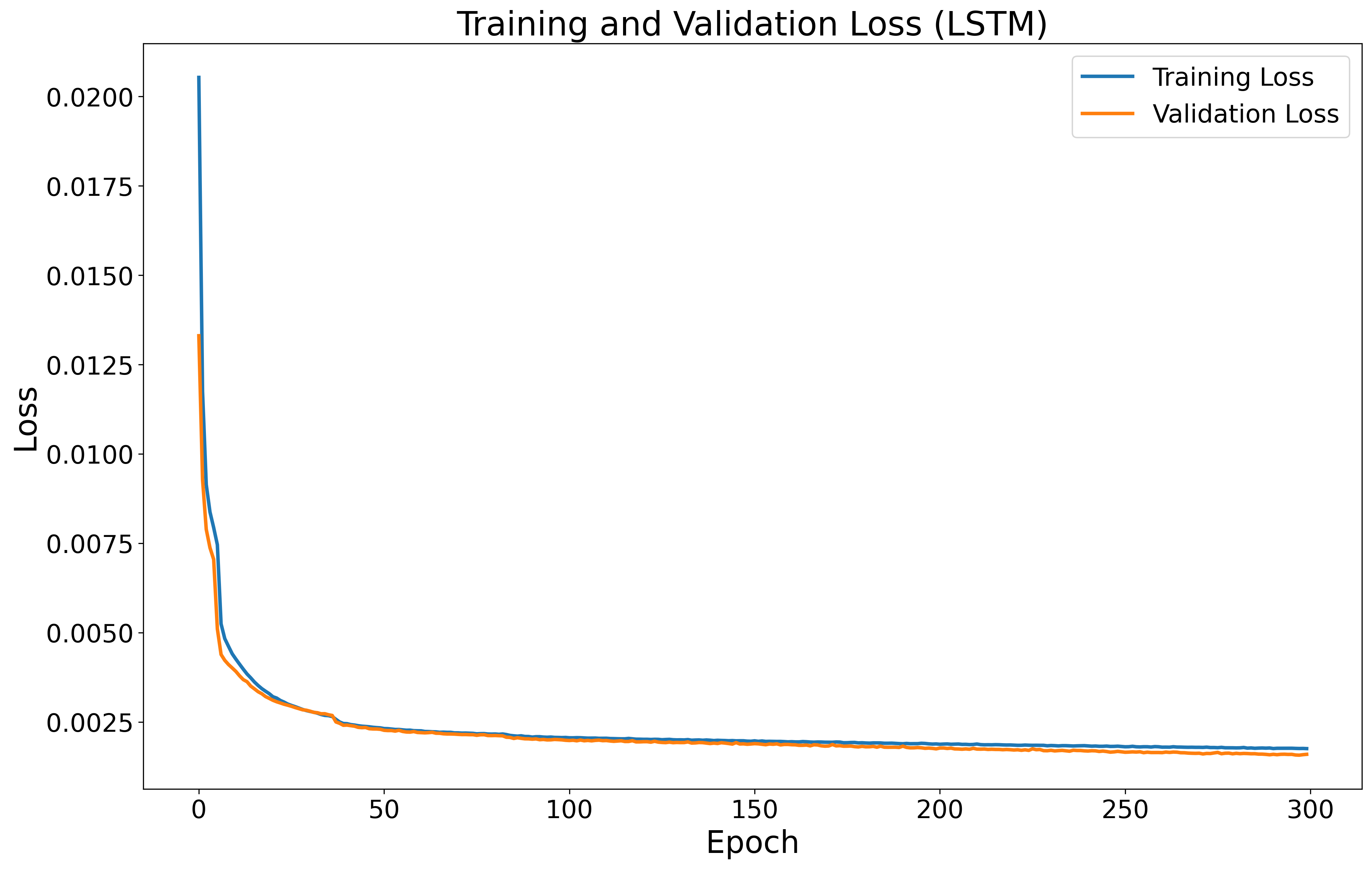}
        \caption{LSTM}
    \end{subfigure}

    \vspace{0.3cm}

    % Second row
    \begin{subfigure}[b]{0.45\textwidth}
        \centering
        \includegraphics[width=\textwidth]{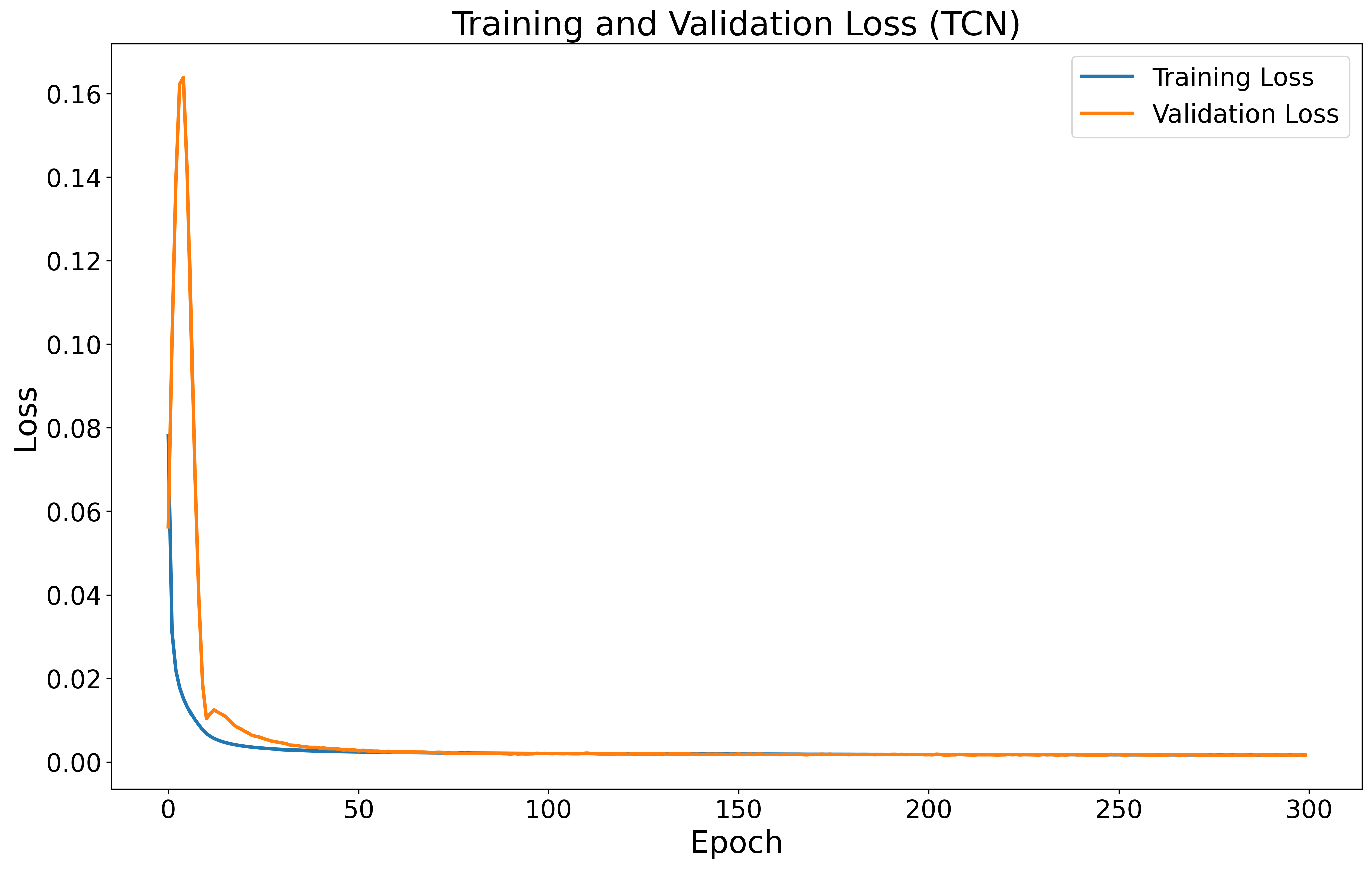}
        \caption{TCN}
    \end{subfigure}
    \hfill
    \begin{subfigure}[b]{0.45\textwidth}
        \centering
        \includegraphics[width=\textwidth]{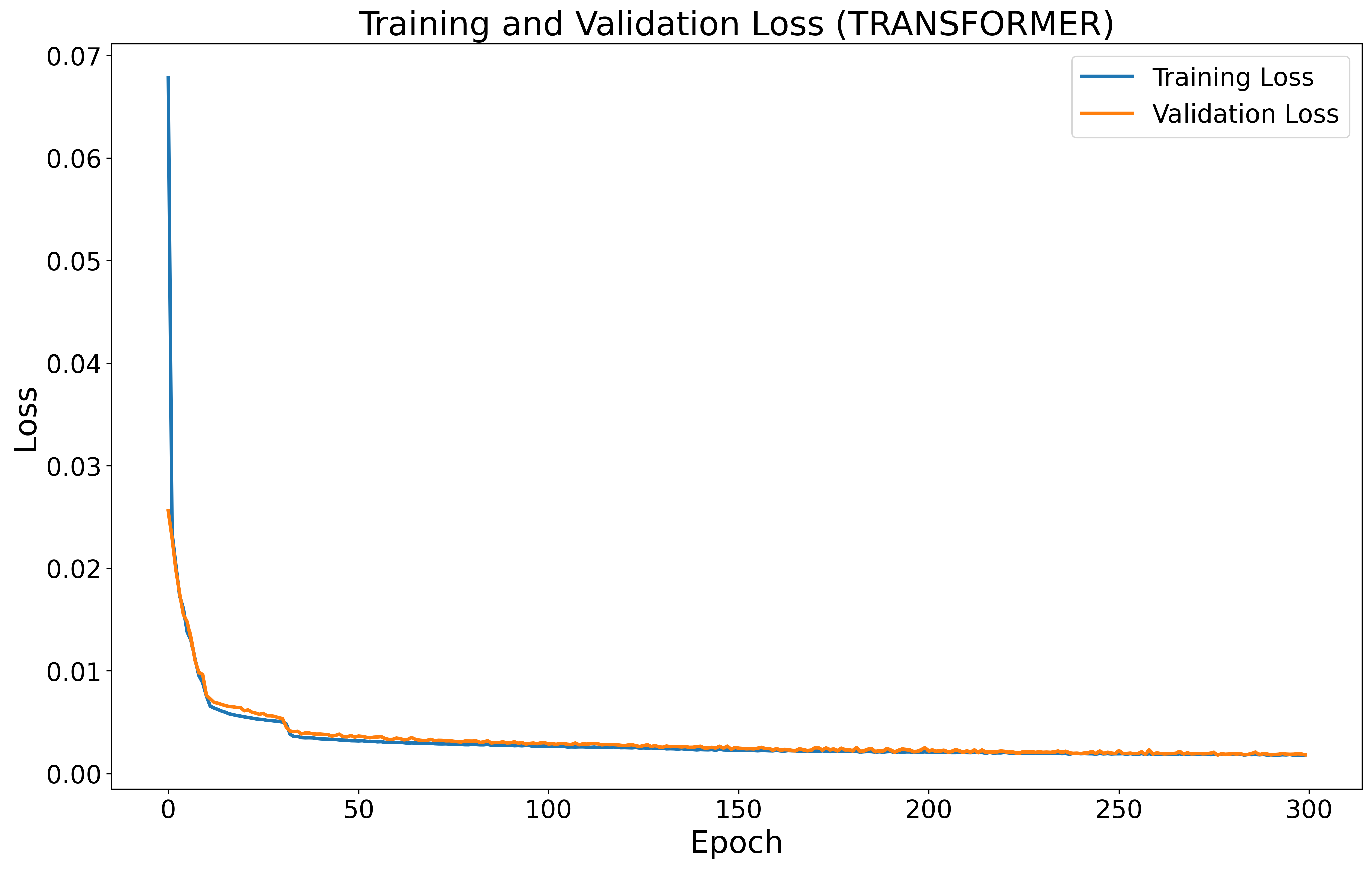}
        \caption{Transformer}
    \end{subfigure}

    \caption{Training and validation loss curves for the four deep learning models (RNN, LSTM, TCN, and Transformer) over 500 epochs. Each plot shows model convergence behavior and generalization performance.}
    \label{fig:loss_subfigures1}
\end{figure}

The four DL models investigated in this work (RNN, LSTM, TCN, and Transformer) successfully learned from the GG time series. As shown in Figure~\ref{fig:loss_subfigures1}, all models converged within 300 epochs, with training and validation losses following parallel trends. This behavior confirms stable training and indicates that the models captured generalizable patterns rather than memorizing the data, enabling reliable forecasting on unseen sequences. Moreover, the nearly identical loss trends between training and validation curves suggest that the models did not suffer from overfitting, highlighting the effectiveness of the regularization strategies employed and the adequacy of the training dataset size. The rapid convergence observed, especially for the LSTM and TCN, indicates that these architectures can efficiently capture temporal dependencies inherent in GG dynamics, where microstructural evolution follows predictable but nonlinear time progressions.\\

Among the tested binning configurations, the setting with 30 bins yielded the best performance across all models achieving their lowest MREs (Table~\ref{tab:mre_models_bins}). Lower bin counts (10 or 20) oversimplified the distribution, while higher values (40 or 50) introduced instability due to sparse intervals. The 30 bin configuration provided the right balance, preserving distributional detail without compromising model stability or generalization. This result reveals that the discretization level plays a crucial role in determining how well the models can represent the continuous evolution of grain size. A too-coarse binning (fewer bins) smooths out critical details such as fine-grain formation and tail evolution, leading to a loss of physical fidelity. Conversely, an overly fine binning creates sparsity that amplifies statistical noise, hindering convergence and increasing prediction variance. The optimal 30-bin setting thus provides a meaningful representation of the microstructural distribution while maintaining computational tractability.\\

\begin{table}[ht]
\centering
\caption{Mean Relative Error (MRE \%) of Different Models Across Numbers of Bins}
% \begin{tabular}{c|c|c|c|c}
\begin{tabular}{c c c c c}
\hline
\textbf{Number of Bins} & \textbf{RNN} & \textbf{LSTM} & \textbf{Transformer} & \textbf{TCN} \\
\hline
10 & 13.40 & 12.20 & 14.00 & 12.90 \\
20 & 13.00 & 11.30 & 13.50 & 12.70 \\
30 & \textbf{12.24} & \textbf{9.40} & \textbf{13.25} & \textbf{12.40} \\
40 & 13.80 & 11.90 & 13.60 & 12.60 \\
50 & 13.20 & 11.70 & 13.40 & 12.80 \\
\hline
\end{tabular}
\label{tab:mre_models_bins}
\end{table}

\begin{figure}[ht]
\centering
\includegraphics[width=0.7\linewidth]{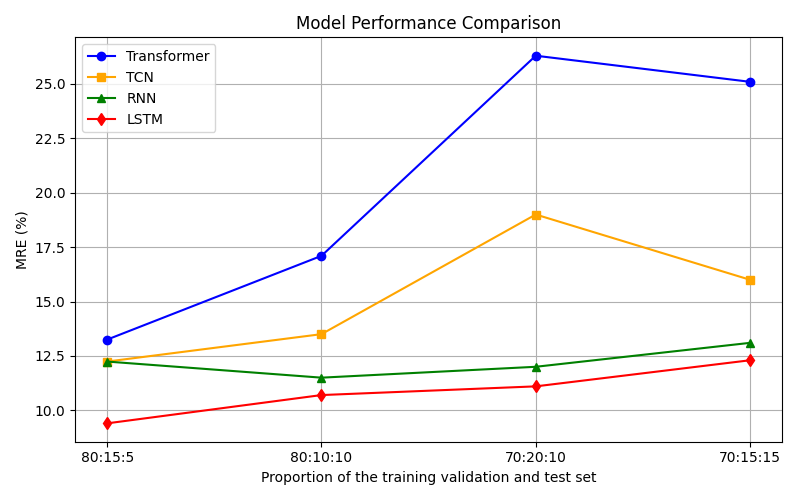}
\caption{Effect of Data Partitioning Strategies on Model Forecasting Accuracy (MRE).}
\label{datasplit}
\end{figure}

For data partitioning, the 80:15:5 training–validation–test split consistently gave the most accurate results (Figure~\ref{datasplit}). Allocating more data to training improved the models’ ability to capture temporal dependencies in grain growth behavior. This trend aligns with expectations in temporal modeling, as a larger training fraction enables the networks to observe a wider range of grain evolution scenarios. The consistent improvements also indicate that the validation and test sets remained representative of the overall data distribution, confirming the robustness of the data split strategy.\\

\begin{figure}[h]
    \centering
    \begin{minipage}[t]{0.48\textwidth}
        \centering
        \includegraphics[width=\linewidth]{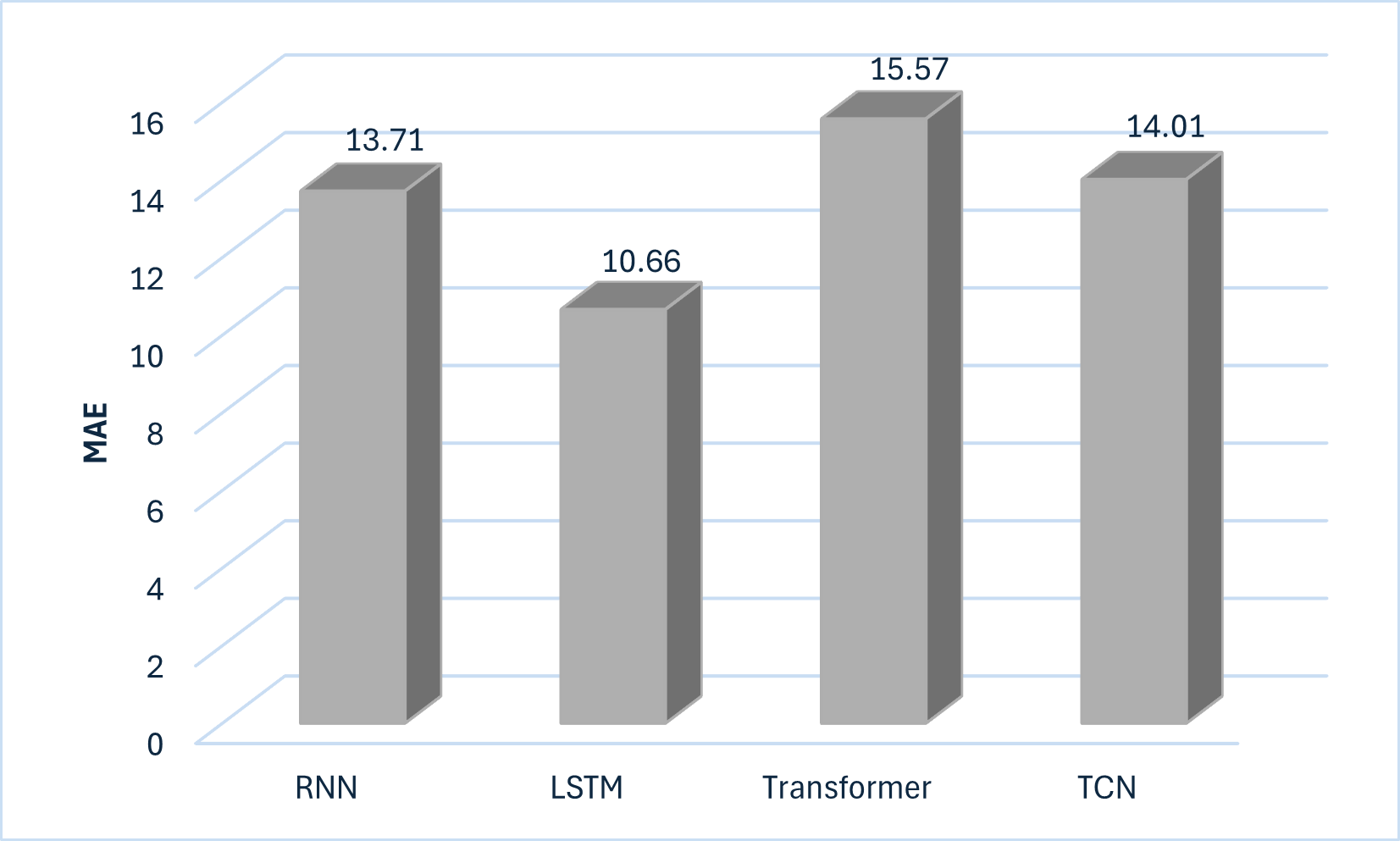}
    \end{minipage}
    \hfill
    \begin{minipage}[t]{0.48\textwidth}
        \centering
        \includegraphics[width=\linewidth]{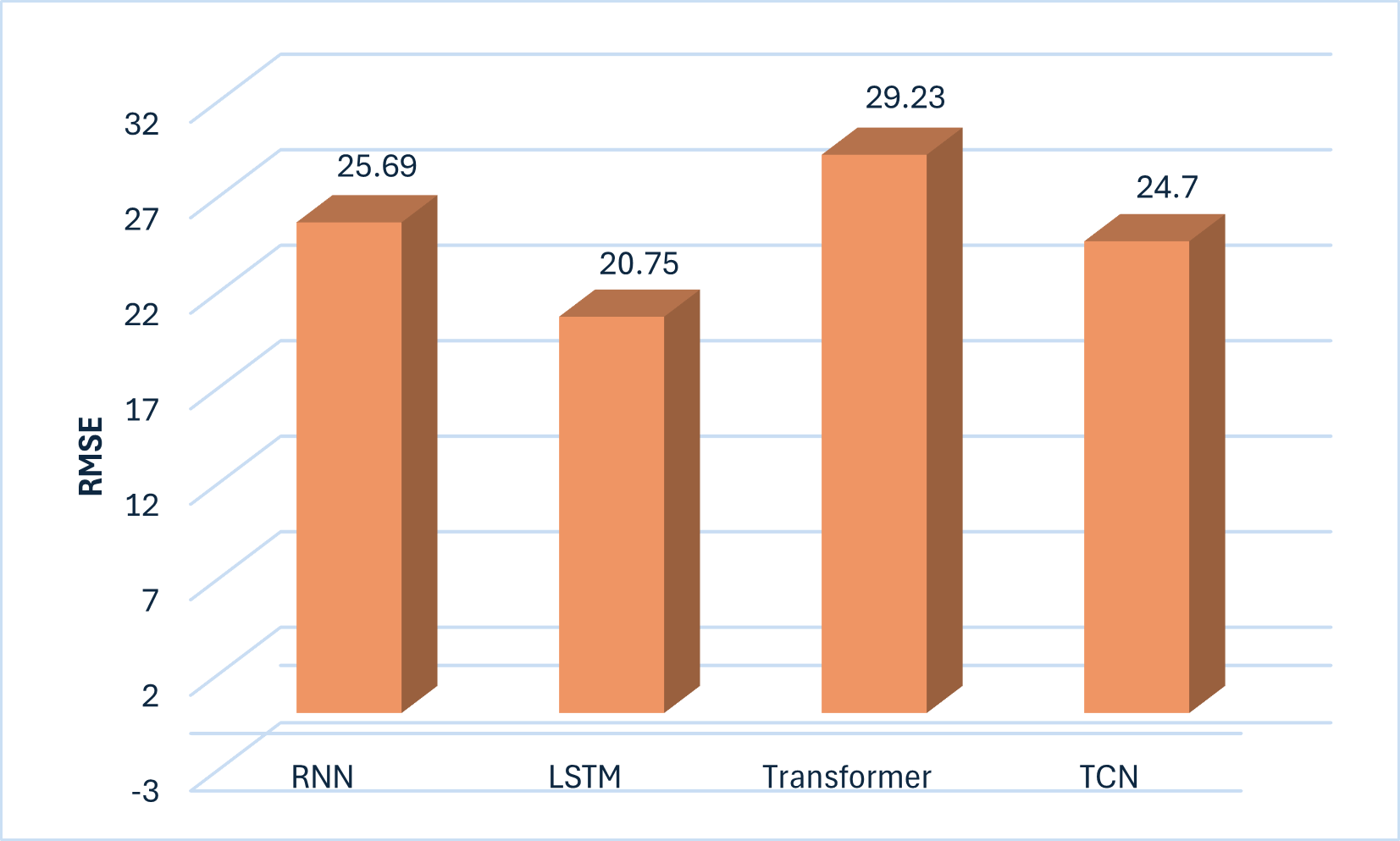}
    \end{minipage}
    \vspace{3mm}
    \begin{minipage}[t]{0.6\textwidth}
        \centering
        \includegraphics[width=\linewidth]{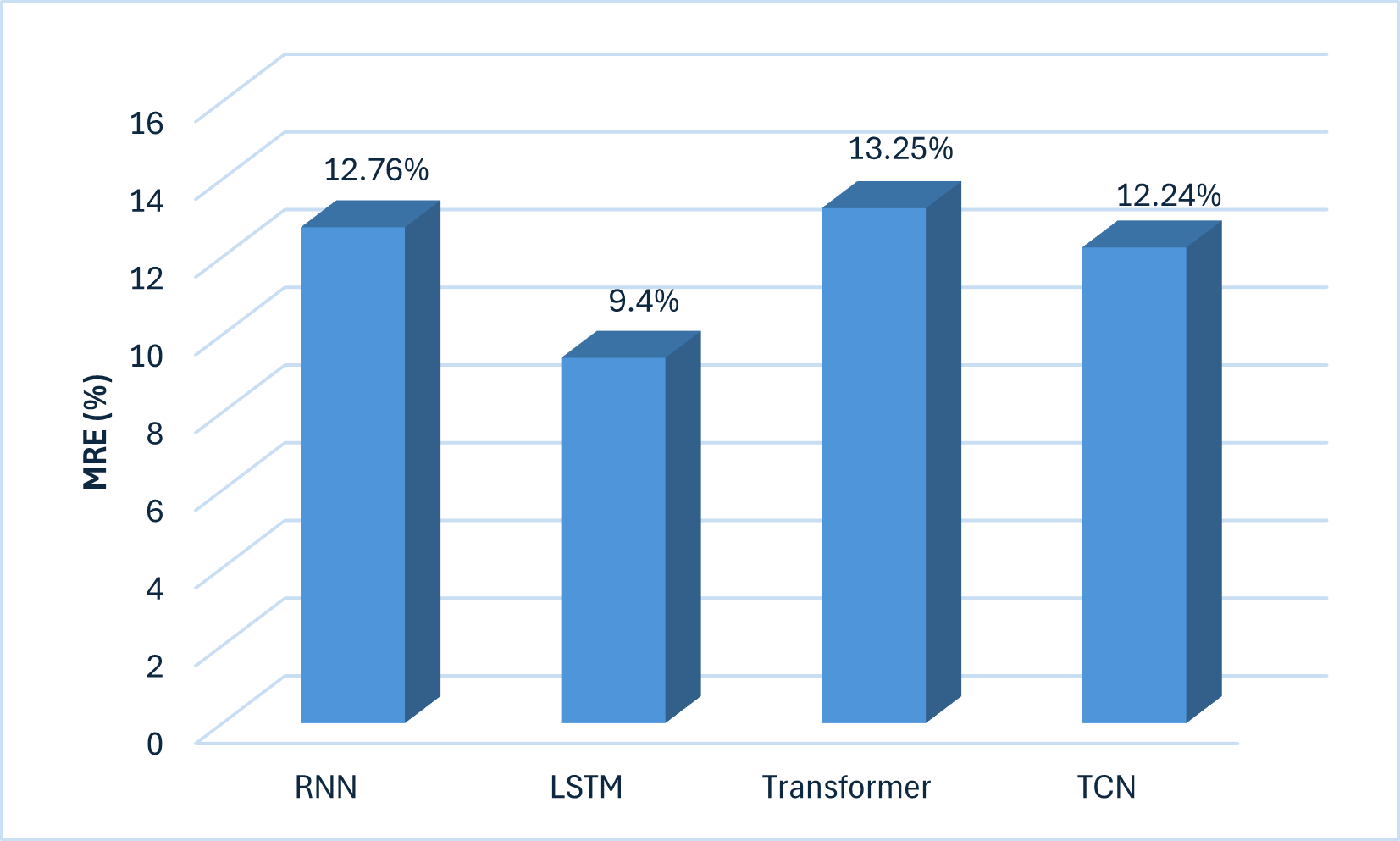}
    \end{minipage}
    \caption{Comparison of prediction performance of different models.}
    \label{fig:model_performance_comparison}
\end{figure}

When comparing predictive performance, the LSTM achieved the lowest errors across all metrics (MAE, RMSE, and MRE), with an average MRE of 9.4\%. This advantage is attributed to its memory cells, which are effective for learning long-term dependencies. The Transformer and TCN also performed well but were more sensitive to hyperparameter choice and dataset size. A summary of these results is presented in Figure~\ref{fig:model_performance_comparison}. 
The superior performance of the LSTM suggests that its gating mechanism effectively captures the sequential relationships between consecutive microstructural states, filtering out irrelevant fluctuations while retaining essential temporal information. In contrast, the Transformer’s self-attention mechanism, though powerful, may have underperformed slightly due to the limited sequence length and smaller dataset size, conditions under which attention-based models require extensive data for optimal generalization. The TCN, while efficient in handling short-range dependencies, appeared less capable of preserving long-term contextual information. Nevertheless, all models achieved mean relative errors below 15\%, demonstrating their strong capability in reproducing realistic grain evolution patterns.\\

\begin{figure}[htp]
% \ContinuedFloat
\centering

\includegraphics[width=0.7\textwidth]{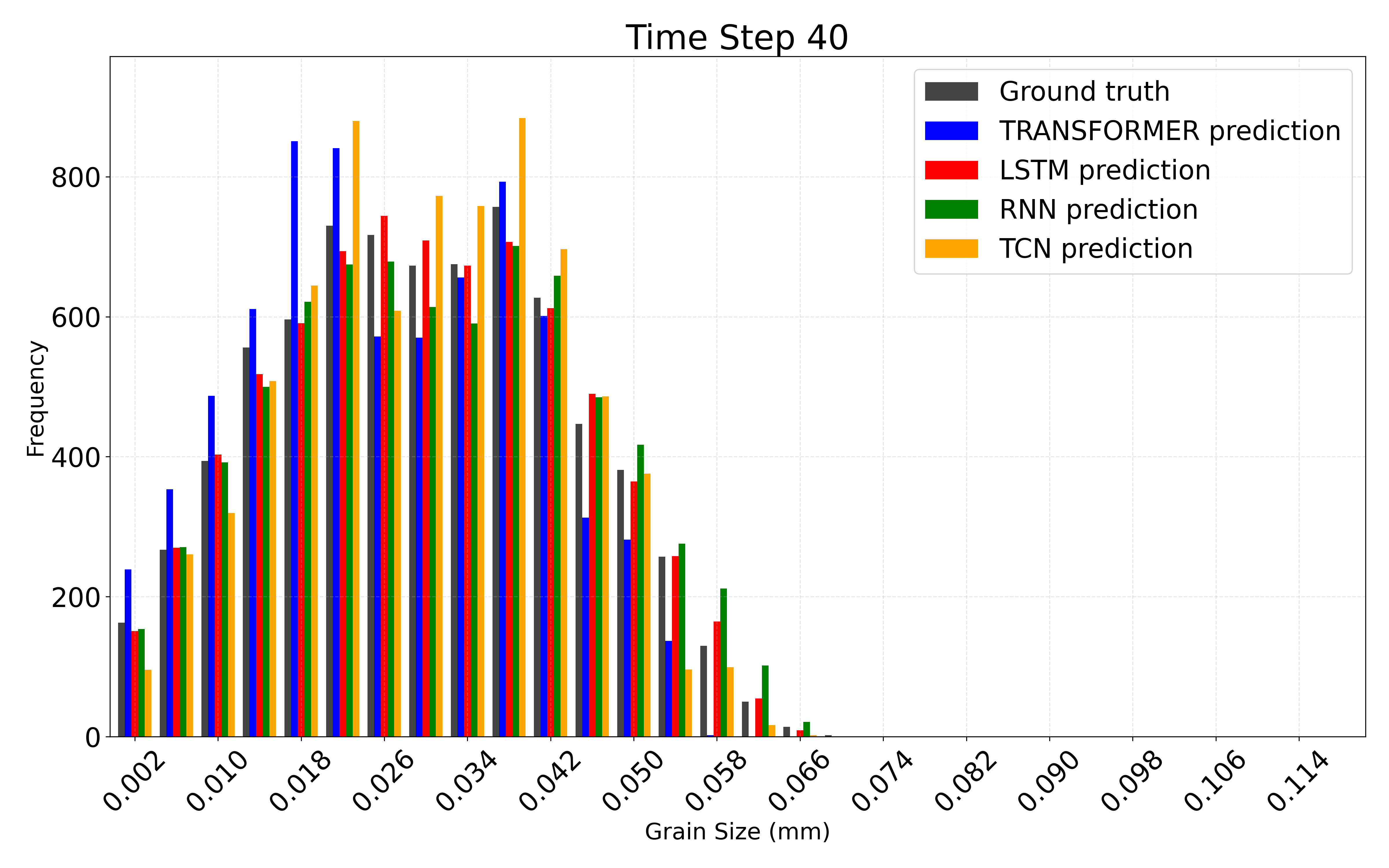}
\caption*{(d) t = 40 min}

\vspace{0.1mm}

\includegraphics[width=0.7\textwidth]{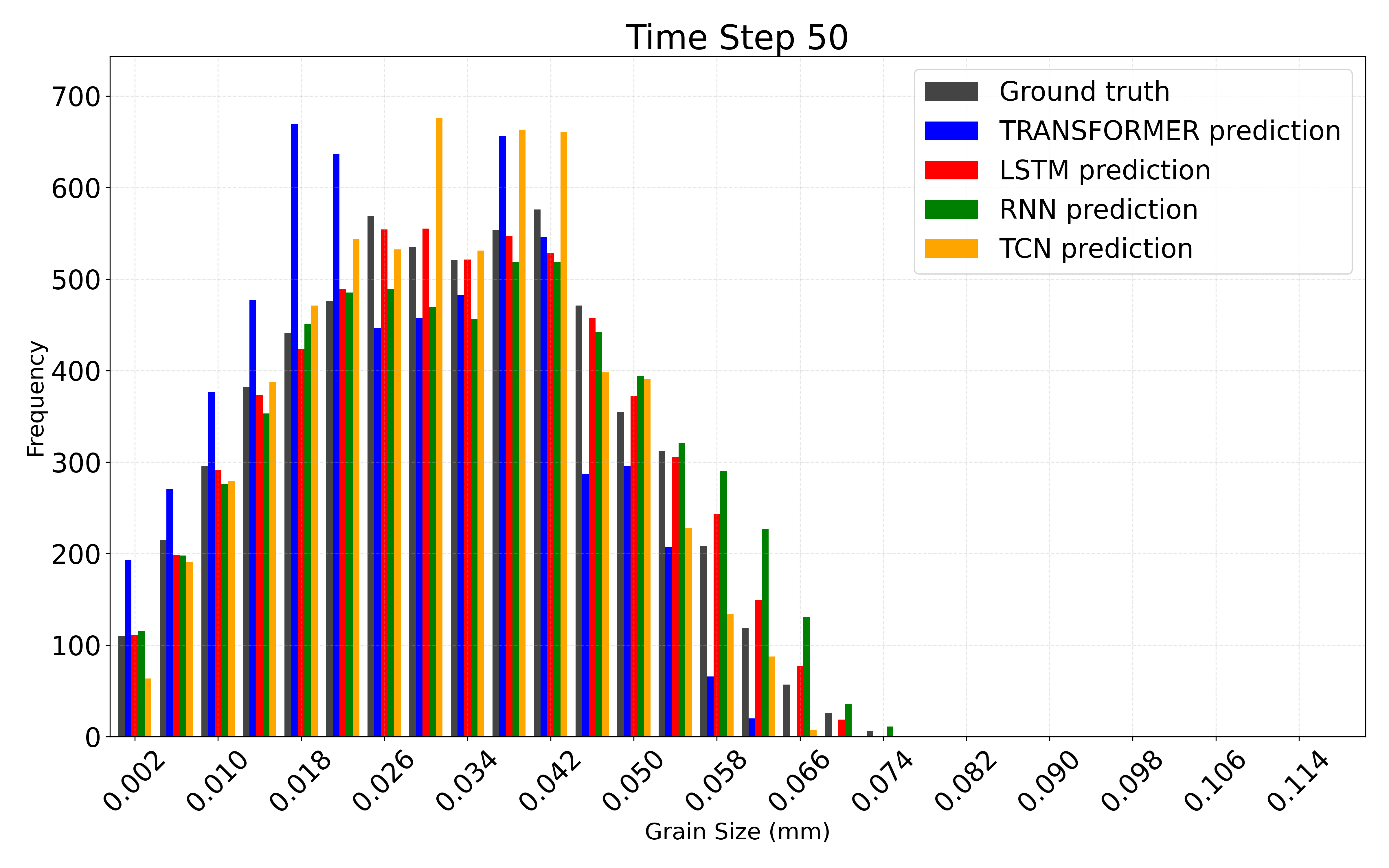}
\caption*{(e) t = 50 min}

\vspace{0.1mm}

\includegraphics[width=0.7\textwidth]{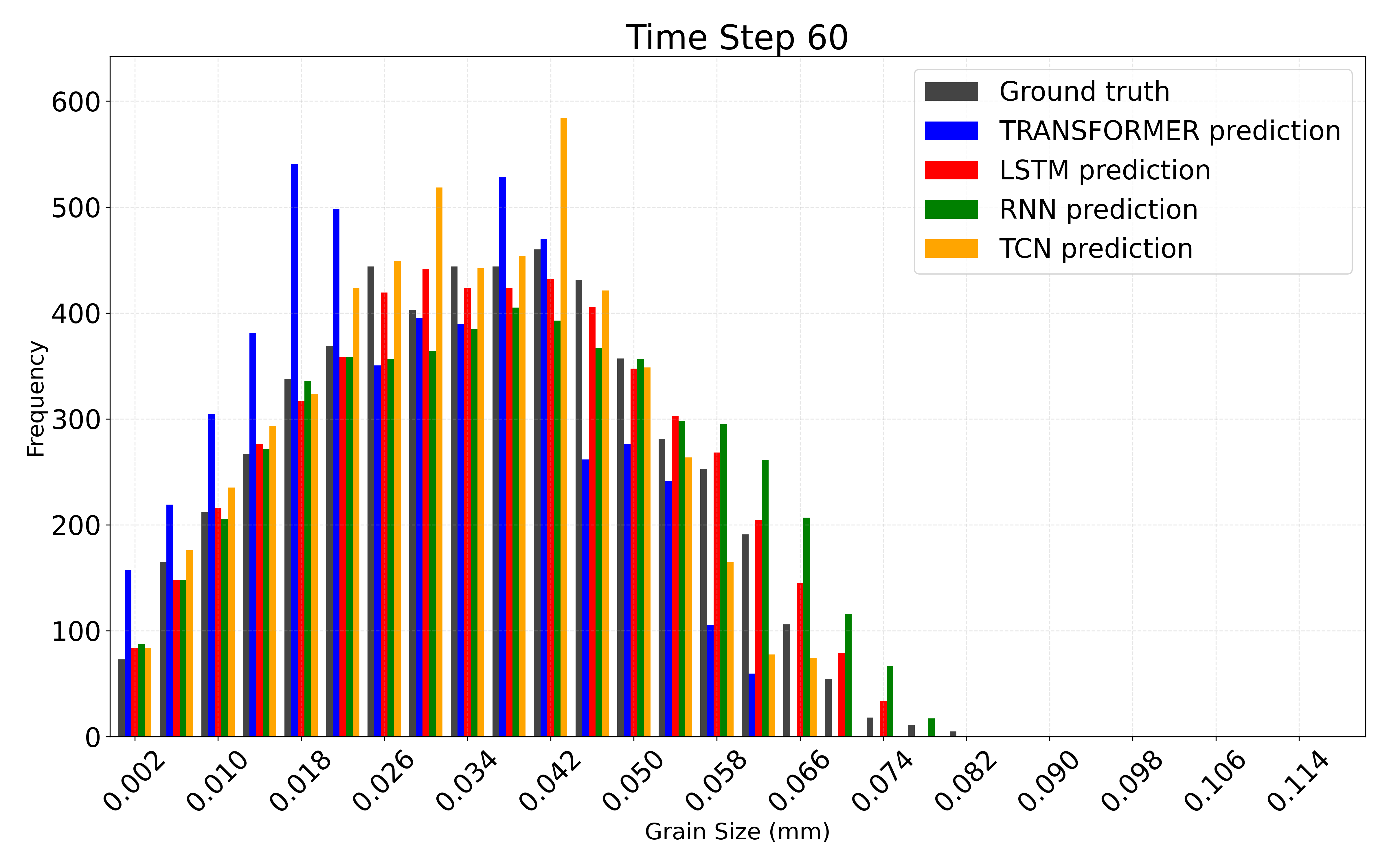}
\caption*{(f) t = 60 min}
\caption{Comparison of Predicted vs. Actual Grain Size Distributions Over Time (5-Minute Input Window Forecasting up to 1 Hour) Using RNN, LSTM, TCN, and Transformer Models. Subfigures (a–f) show predictions at different times for a single simulation case.}
\label{fig:six_images_comparison2}
\end{figure}

To further assess model performance, their ability to reproduce the temporal evolution of grain size distributions were examined. Figure~\ref{fig:six_images_comparison2} illustrates predictions for a representative case at successive time steps. The transition from an initially Gaussian-like profile to a lognormal distribution is well captured, with the LSTM closely tracking the TRM reference. Notably, the model reproduces both peak flattening and tail broadening, indicating its capacity to forecast not only the mean grain size but also the evolution of the entire distribution. This agreement with physical grain growth behavior, where coarsening naturally leads to broader, skewed size distributions, demonstrates that the model learns the underlying kinetics rather than simply fitting numerical trends. The accurate reproduction of the long-tail region further confirms that the model captures the slow growth of larger grains, which dominate the later stages of the process. Such fidelity is crucial for predictive modeling in materials design, as it ensures that the predicted distribution reflects the correct statistical and physical behavior of the system.\\

\begin{figure}[htp]
\centering

\includegraphics[width=0.7\textwidth]{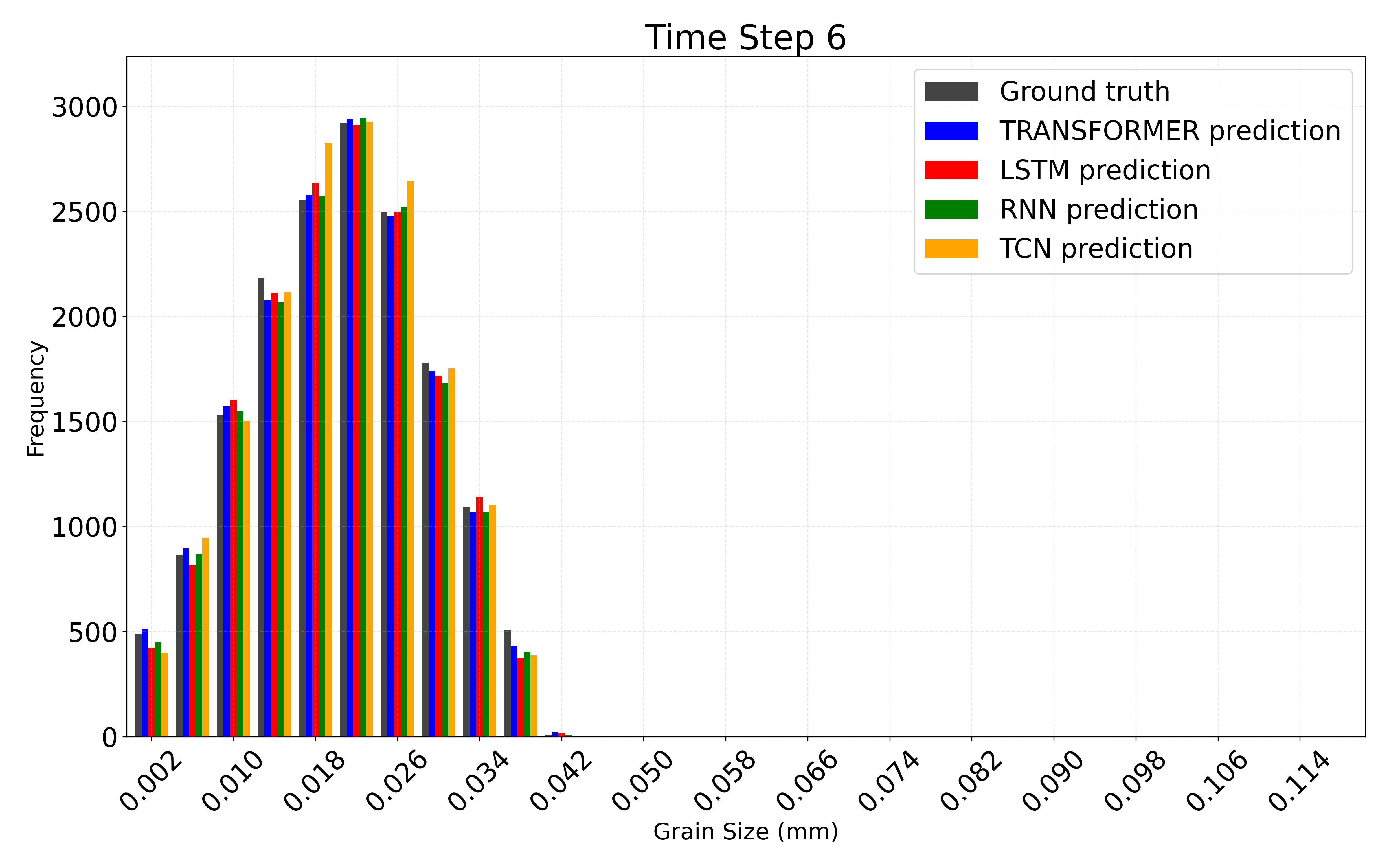}
\caption*{(a) t = 6 min}

\vspace{0.1mm}

\includegraphics[width=0.7\textwidth]{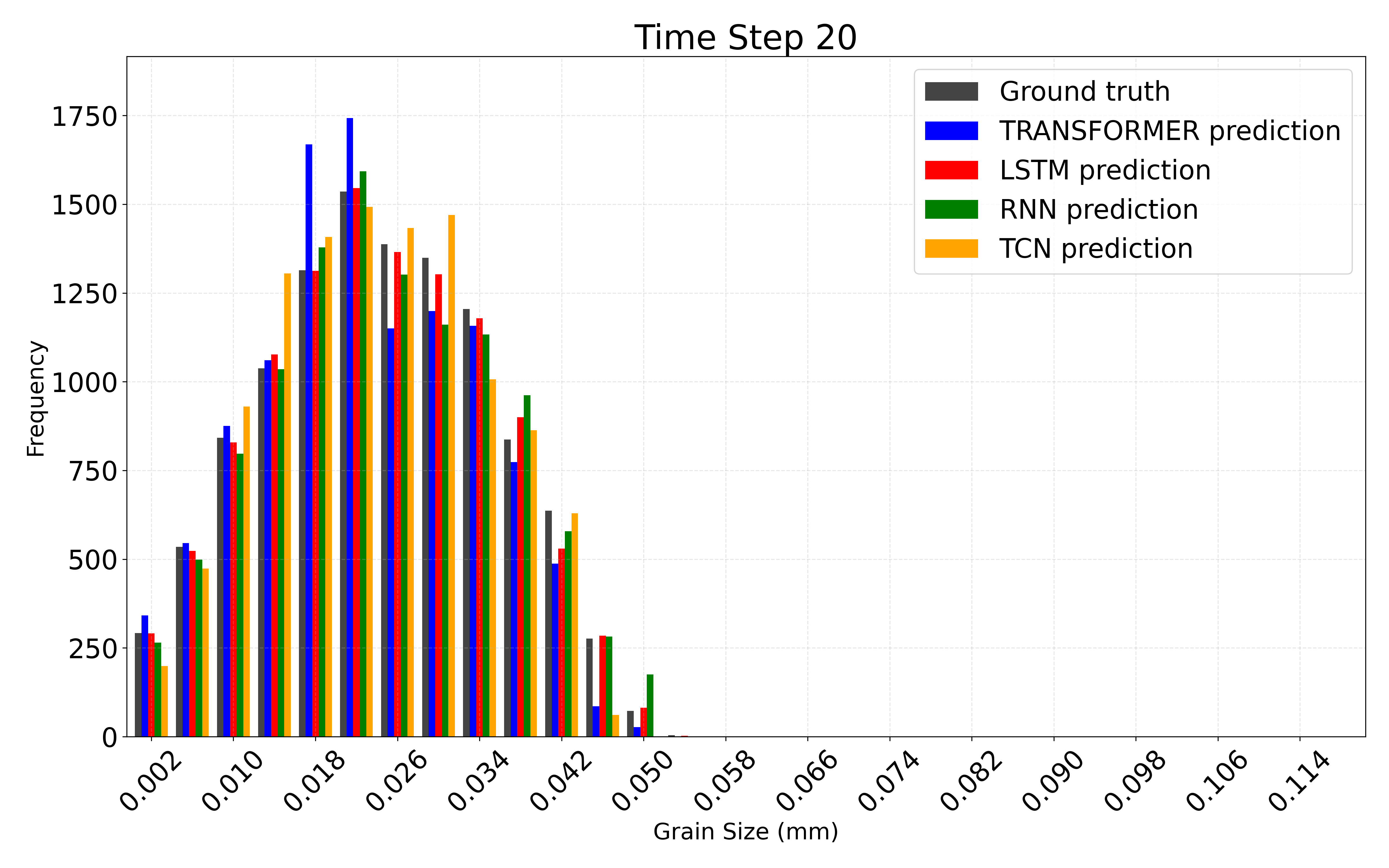}
\caption*{(b) t = 20 min}

\vspace{0.1mm}

\includegraphics[width=0.7\textwidth]{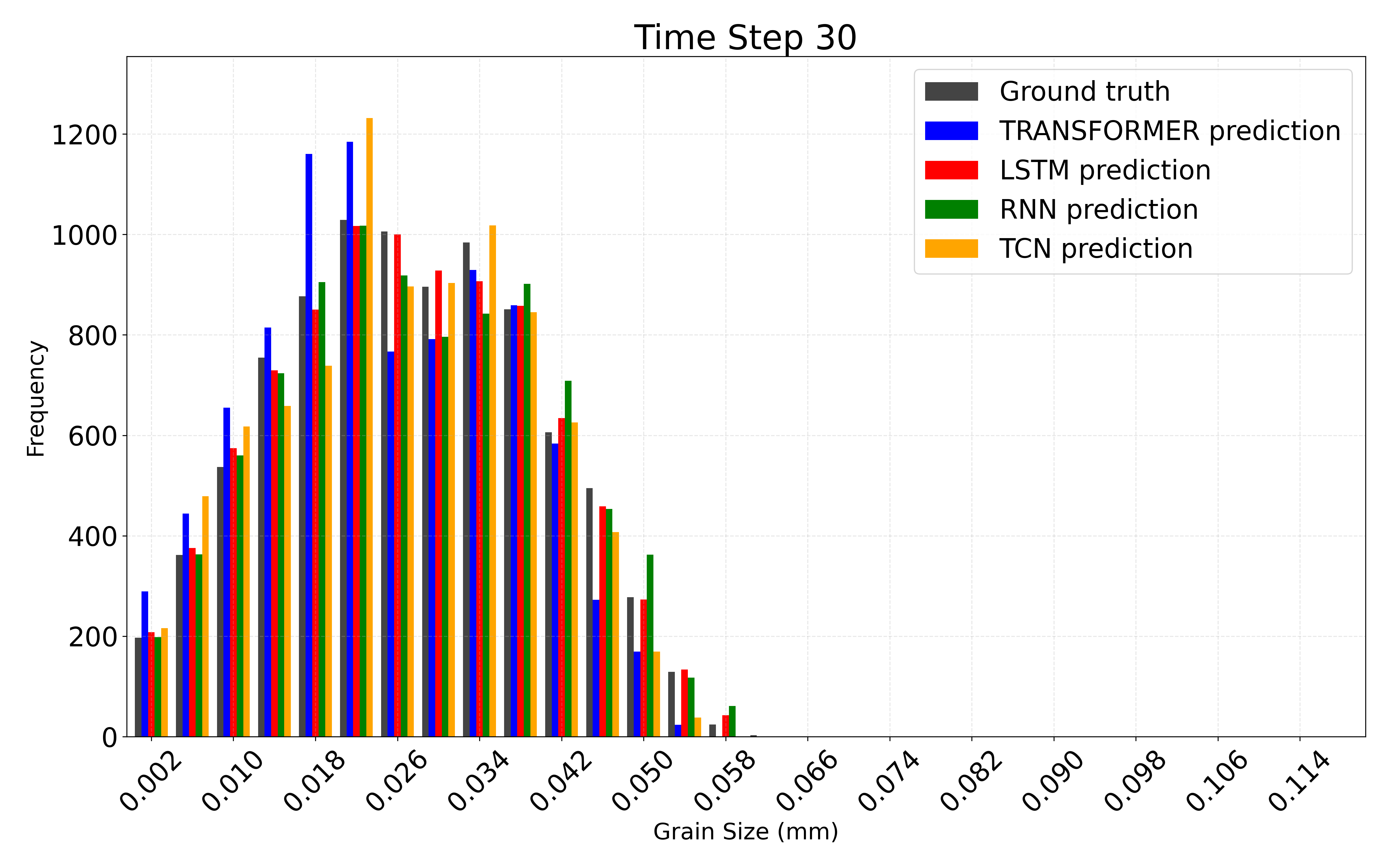}
\caption*{(c) t = 30 min}

\caption*{}
\label{fig:six_images_comparison}
\end{figure}

\begin{figure}[htp]
\centering

% ------- Column headers -------
\makebox[0.32\textwidth][c]{\text{(A)}}
\hfill
\makebox[0.32\textwidth][c]{\text{(B)}}
\hfill
\makebox[0.32\textwidth][c]{\text{(C)}}

\vspace{0.5mm}

% ------- Row: t = 6 min -------

% \includegraphics[width=0.32\textwidth]{sim1.png}
% \hfill
% \includegraphics[width=0.32\textwidth]{sim11.png}
% \hfill
% \includegraphics[width=0.32\textwidth]{sim111.png}
% \text{t = 6 min}
% \vspace{0.5mm}

\includegraphics[width=0.32\textwidth]{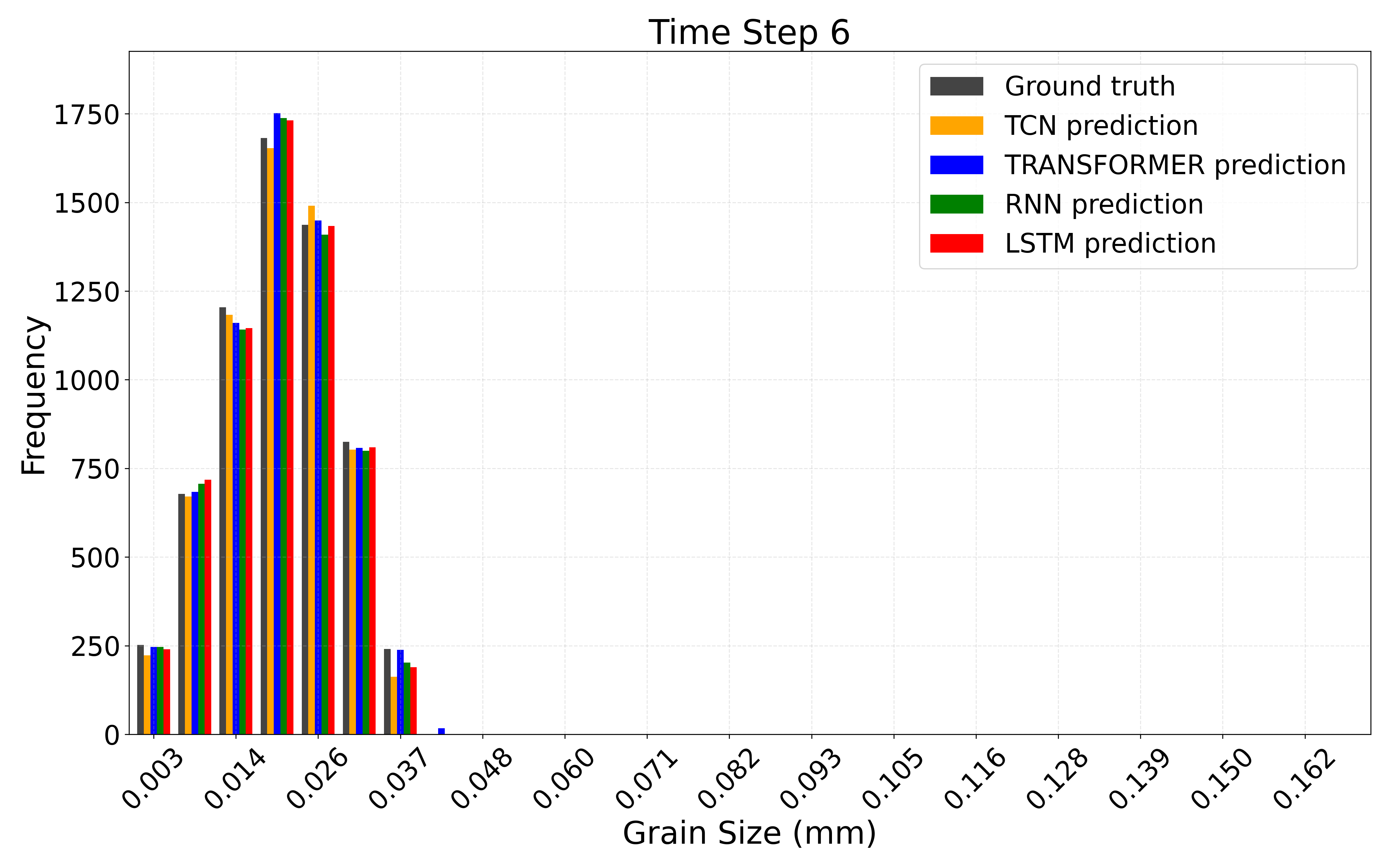}
\hfill
\includegraphics[width=0.32\textwidth]{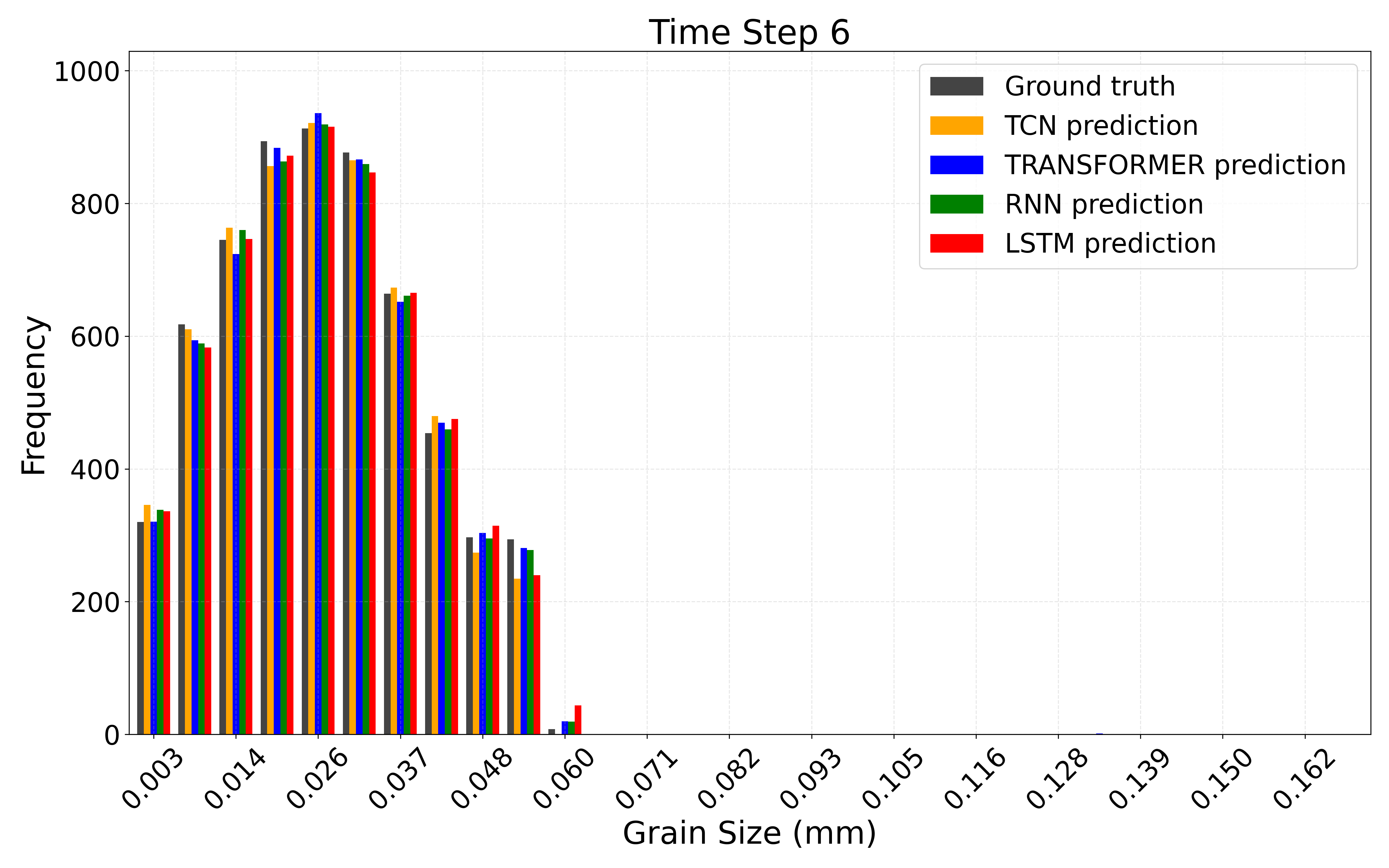}
\hfill
\includegraphics[width=0.32\textwidth]{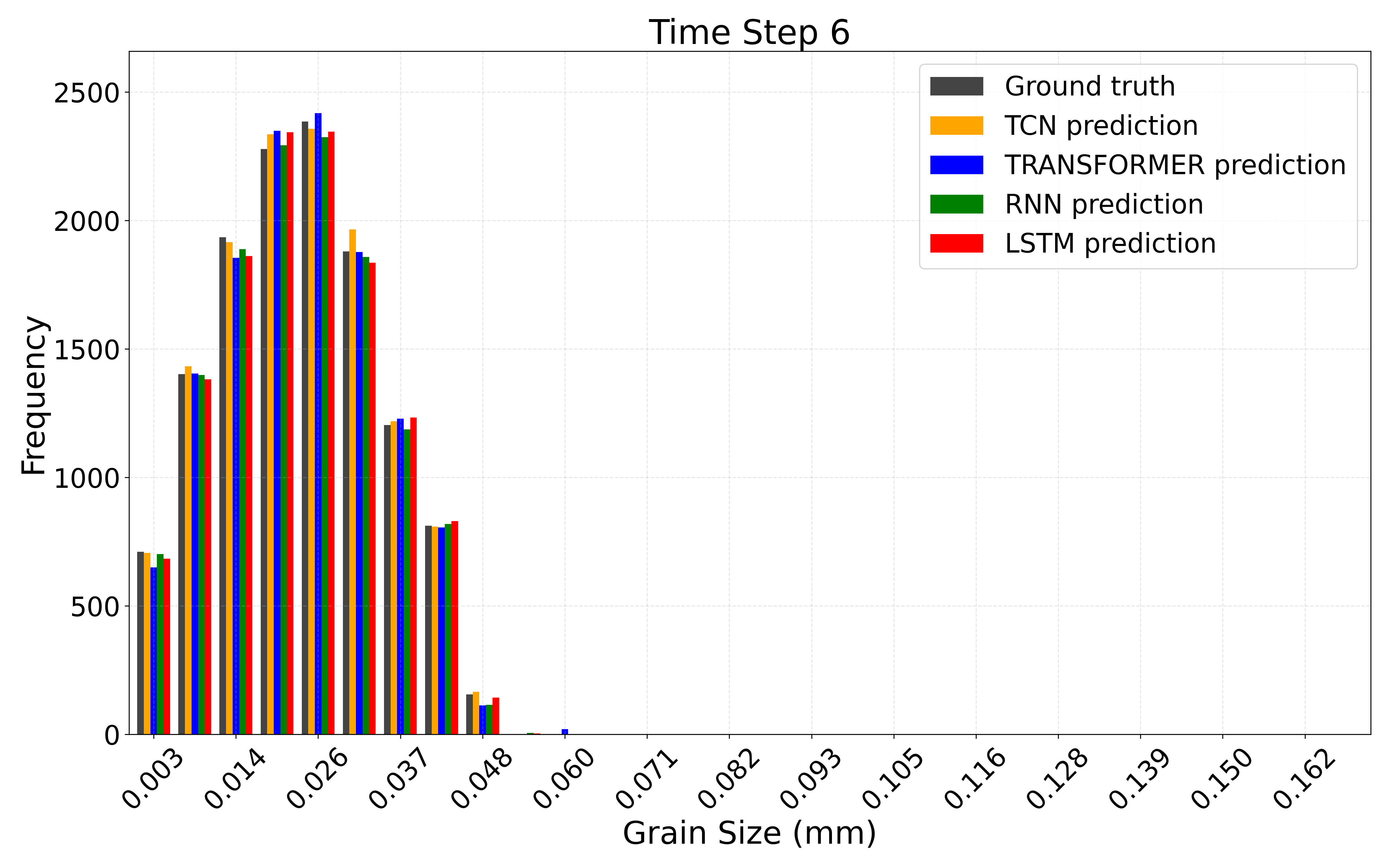}
\text{t = 6 min}
\vspace{0.5mm}
% ------- Row: t = 20 min -------

\includegraphics[width=0.32\textwidth]{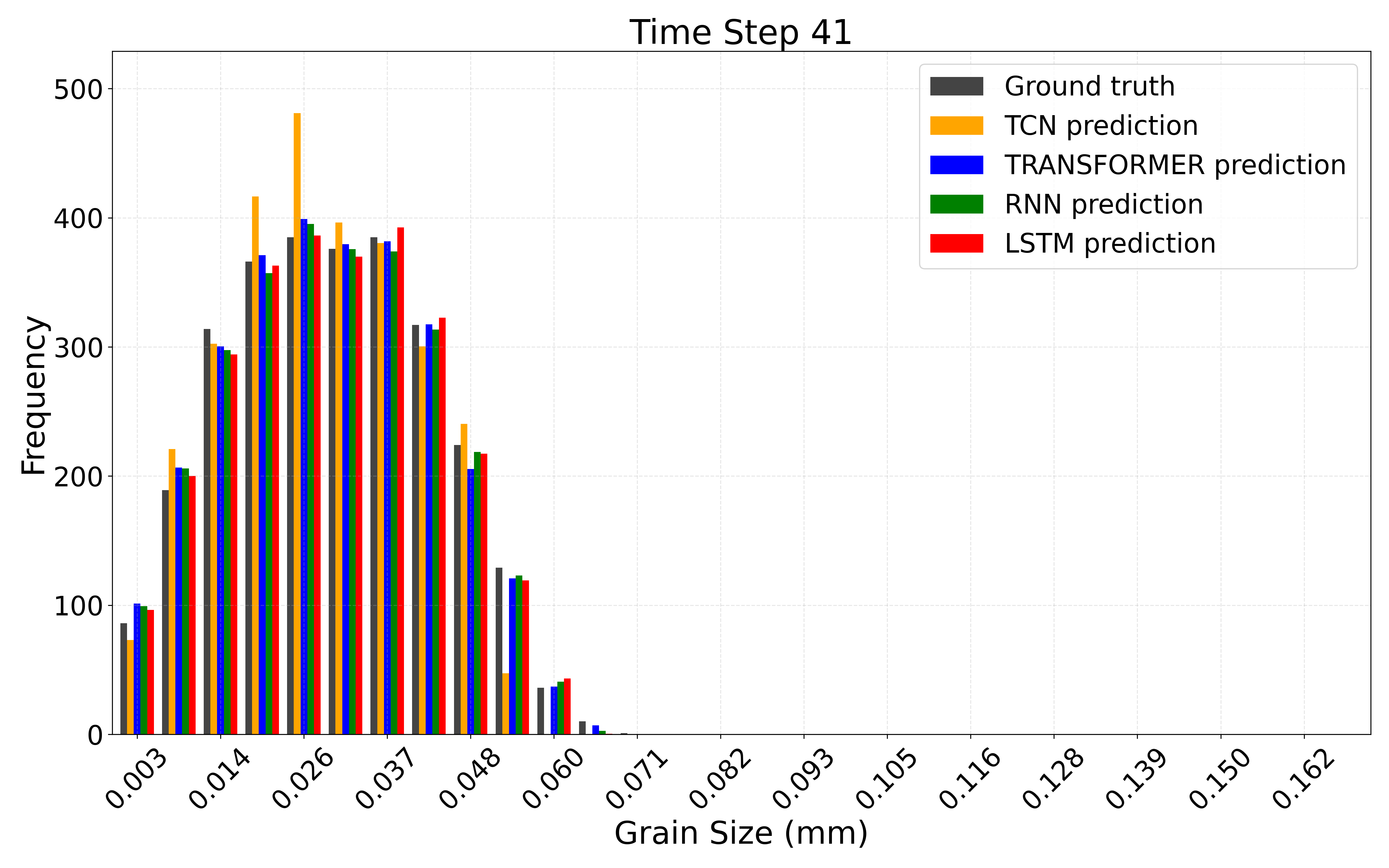}
\hfill
\includegraphics[width=0.32\textwidth]{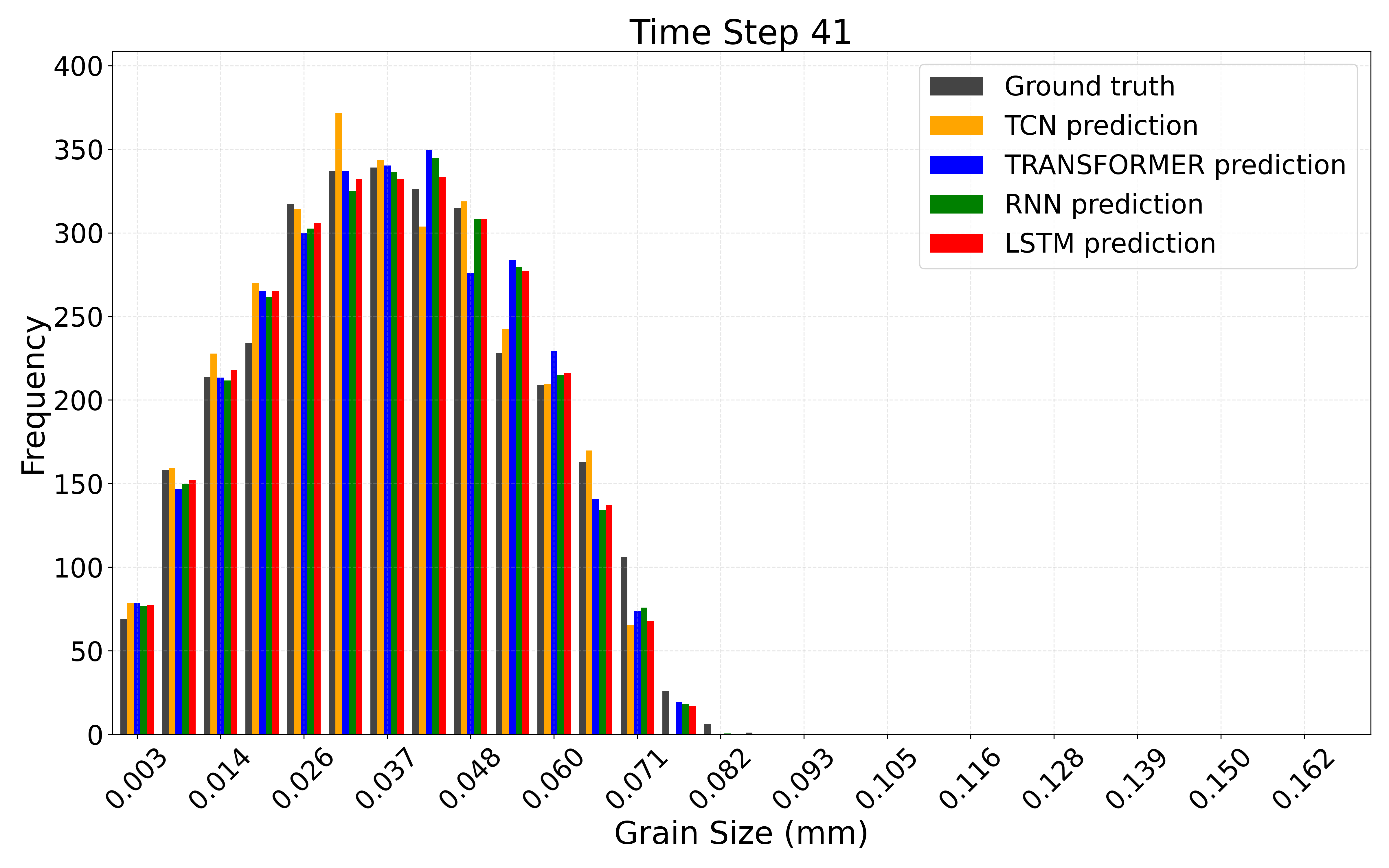}
\hfill
\includegraphics[width=0.32\textwidth]{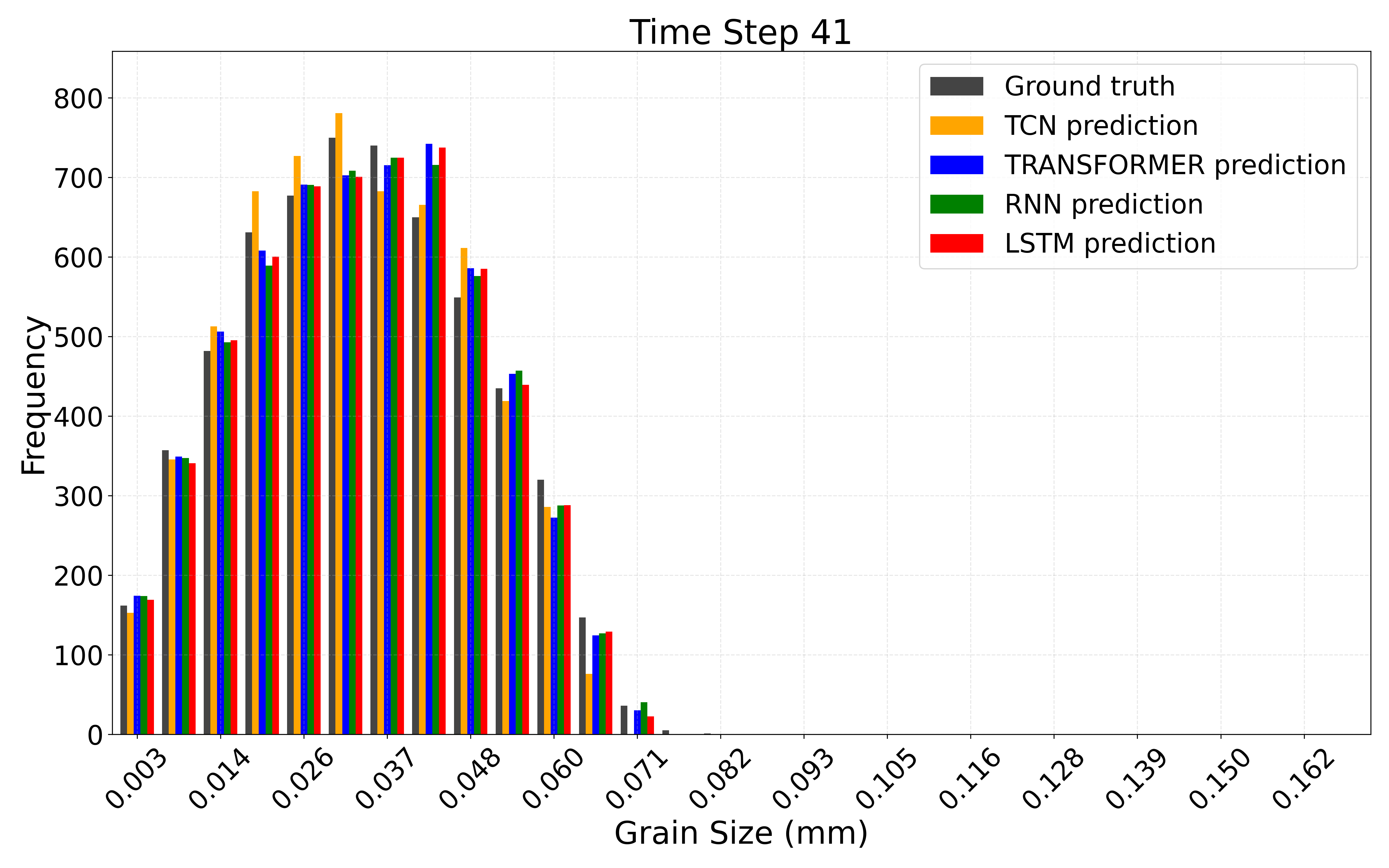}
\text{t = 40 min}
\vspace{0.5mm}

% ------- Row: t = 30 min -------

\includegraphics[width=0.32\textwidth]{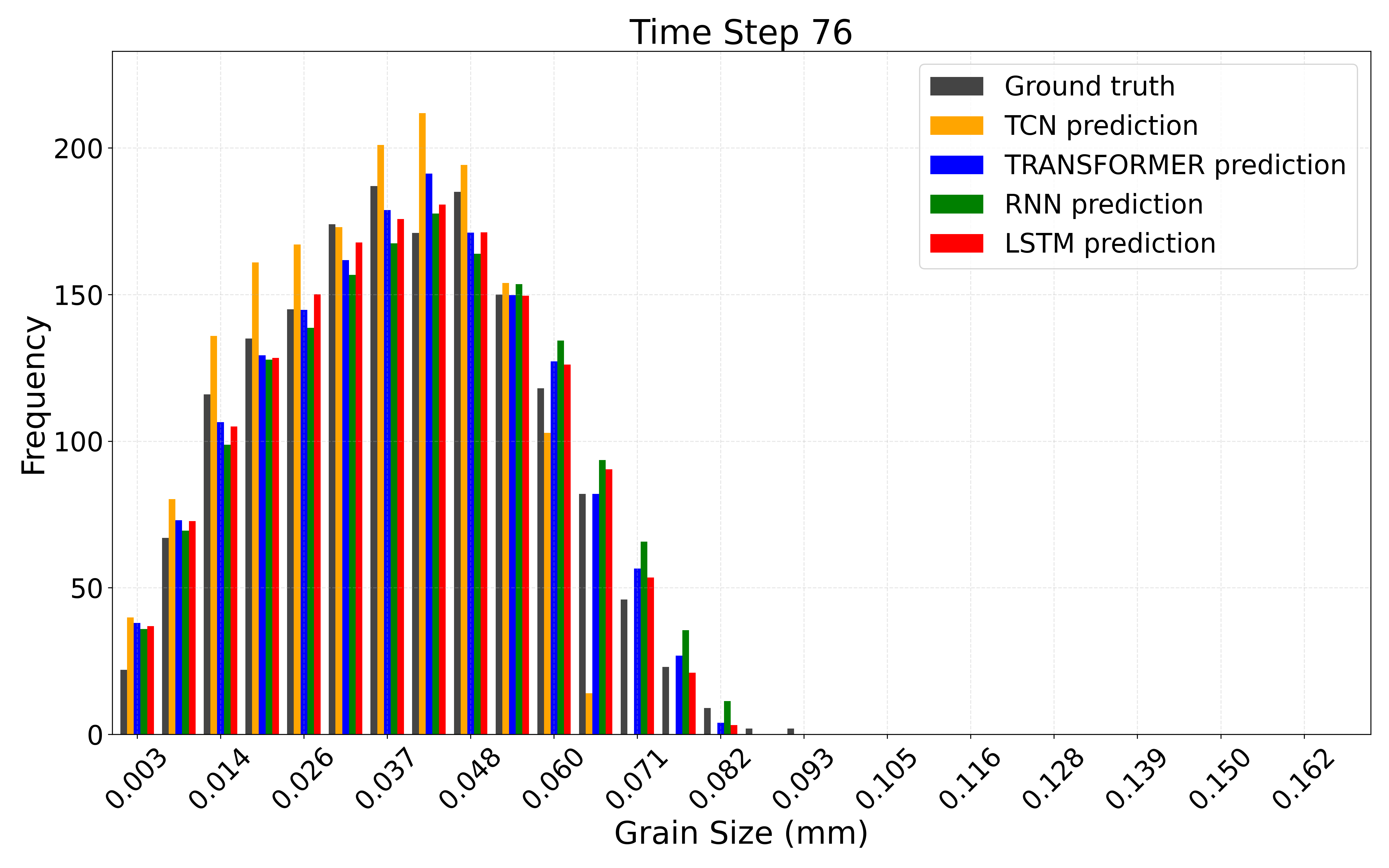}
\hfill
\includegraphics[width=0.32\textwidth]{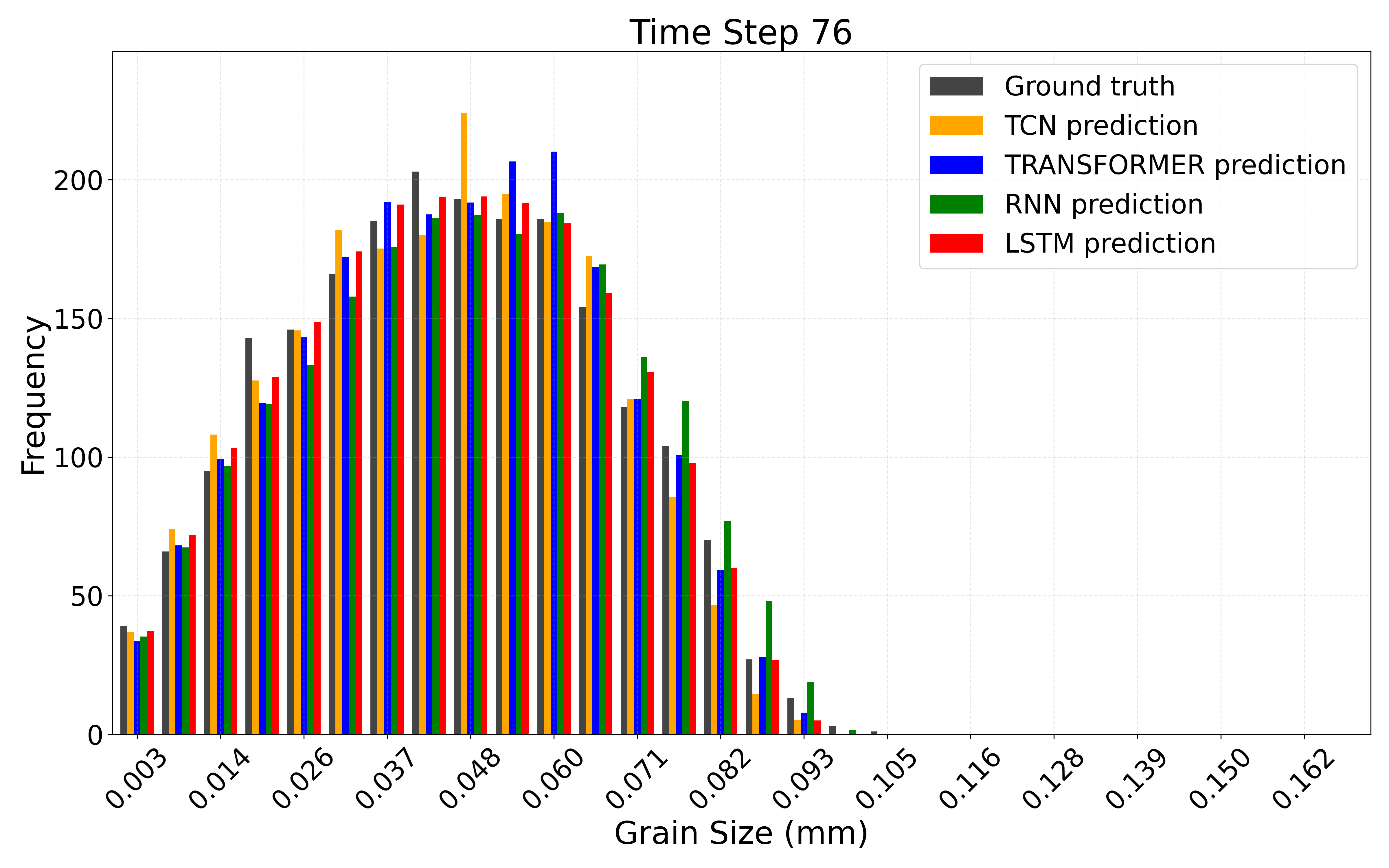}
\hfill
\includegraphics[width=0.32\textwidth]{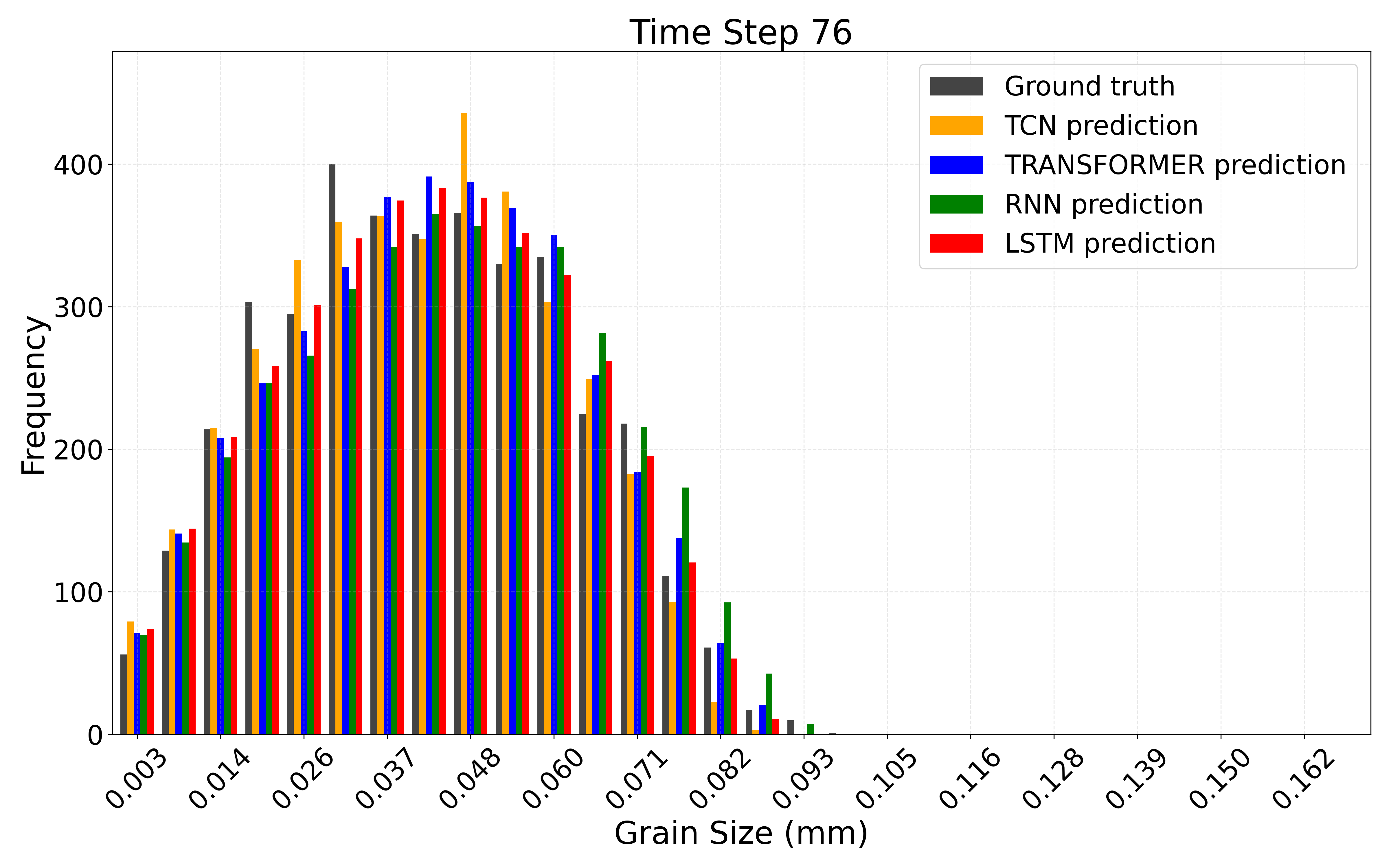}
\text{t = 75 min}\\

\vspace{0.5mm}

% ------- Row: t = 40 min -------

\includegraphics[width=0.32\textwidth]{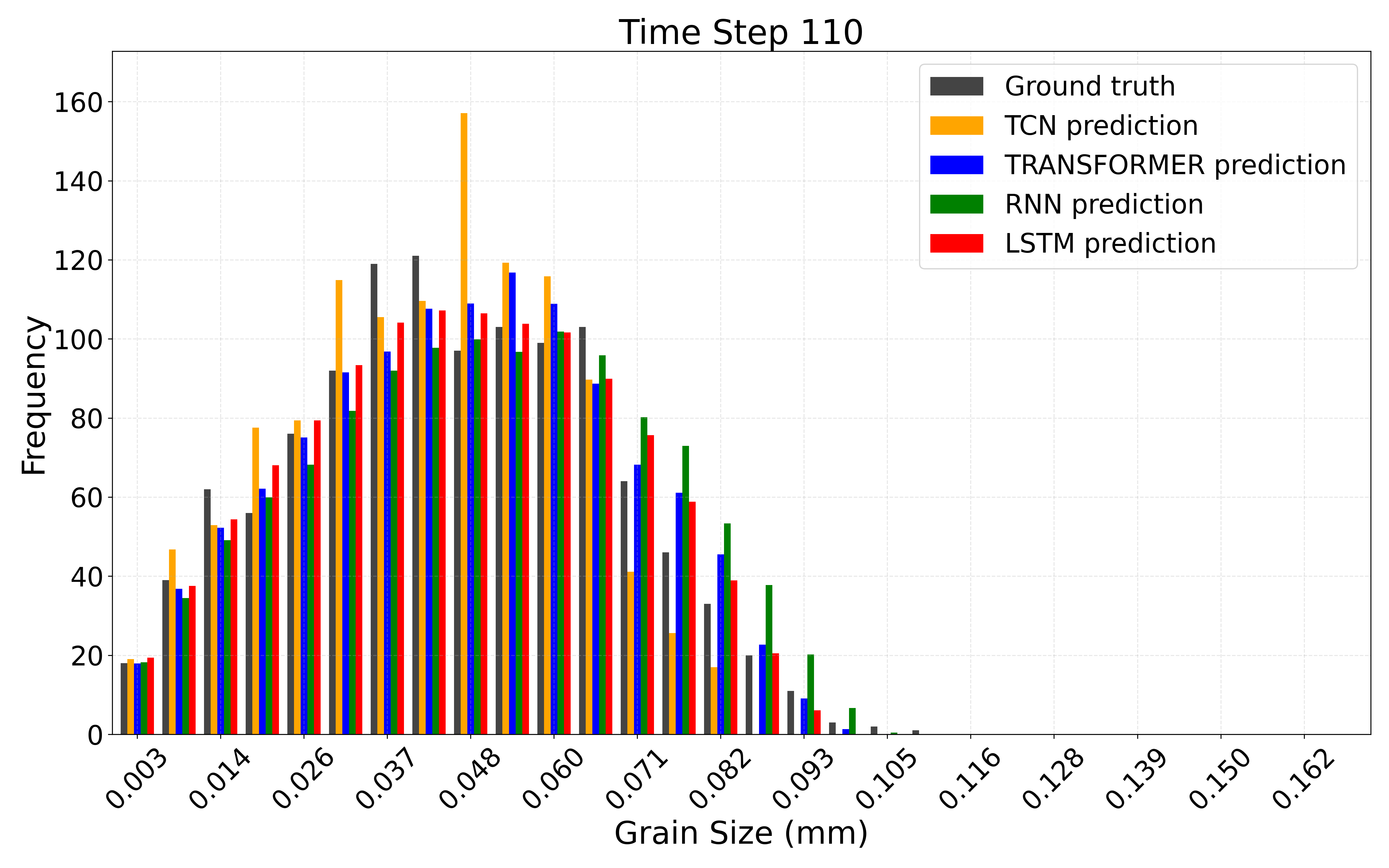}
\hfill
\includegraphics[width=0.32\textwidth]{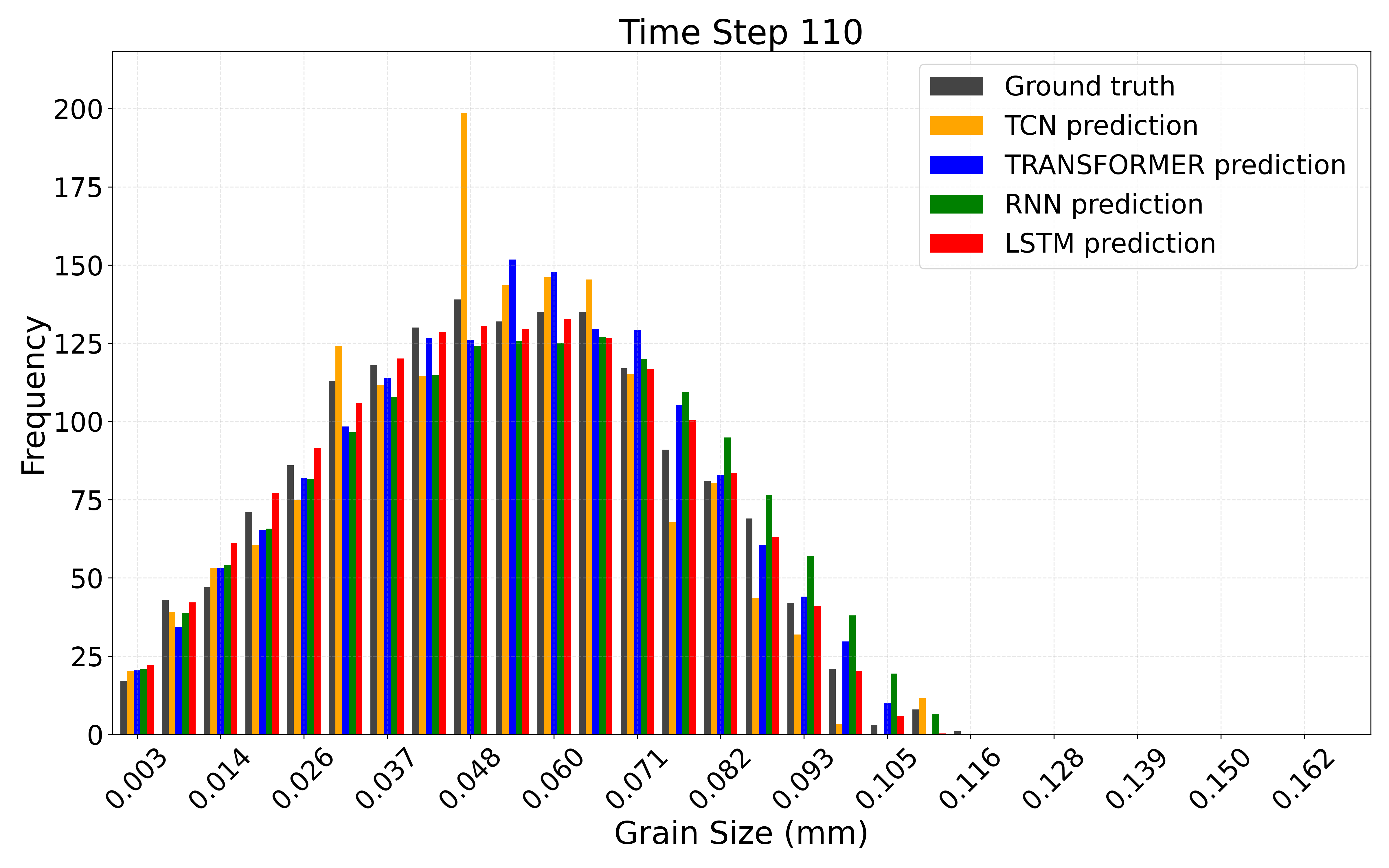}
\hfill
\includegraphics[width=0.32\textwidth]{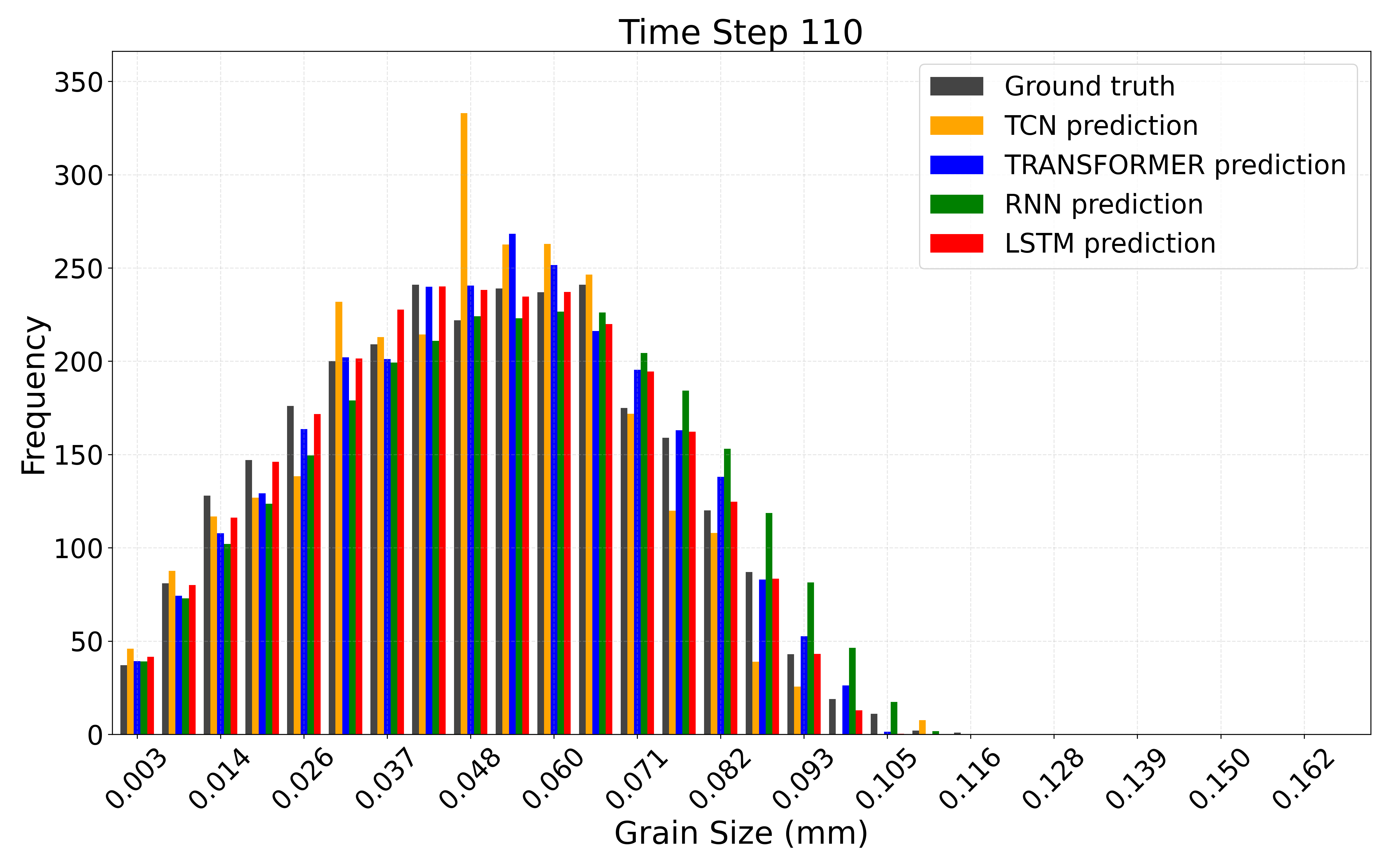}
\text{t = 110 min} \\

\vspace{0.5mm}

% ------- Row: t = 50 min -------

\includegraphics[width=0.32\textwidth]{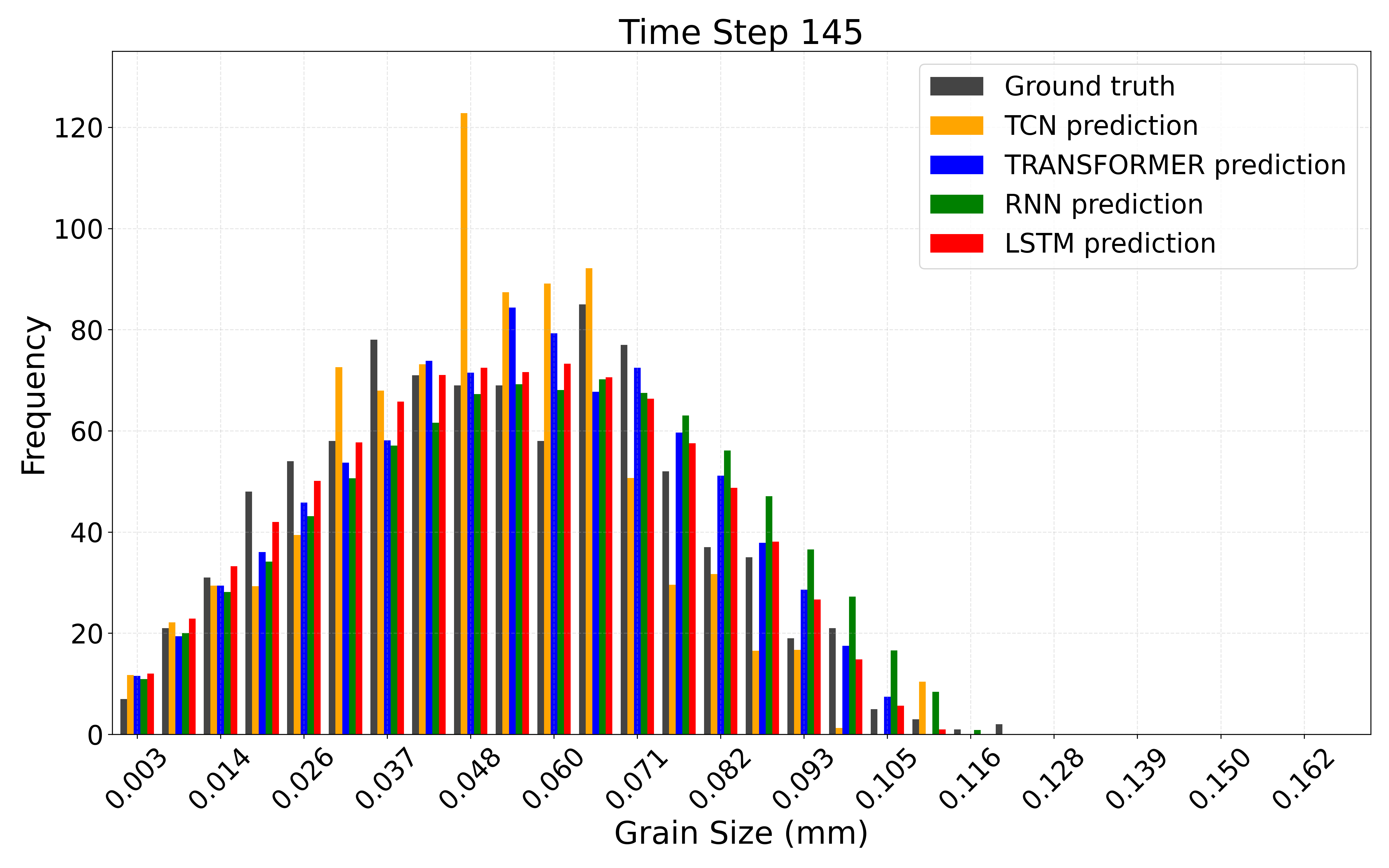}
\hfill
\includegraphics[width=0.32\textwidth]{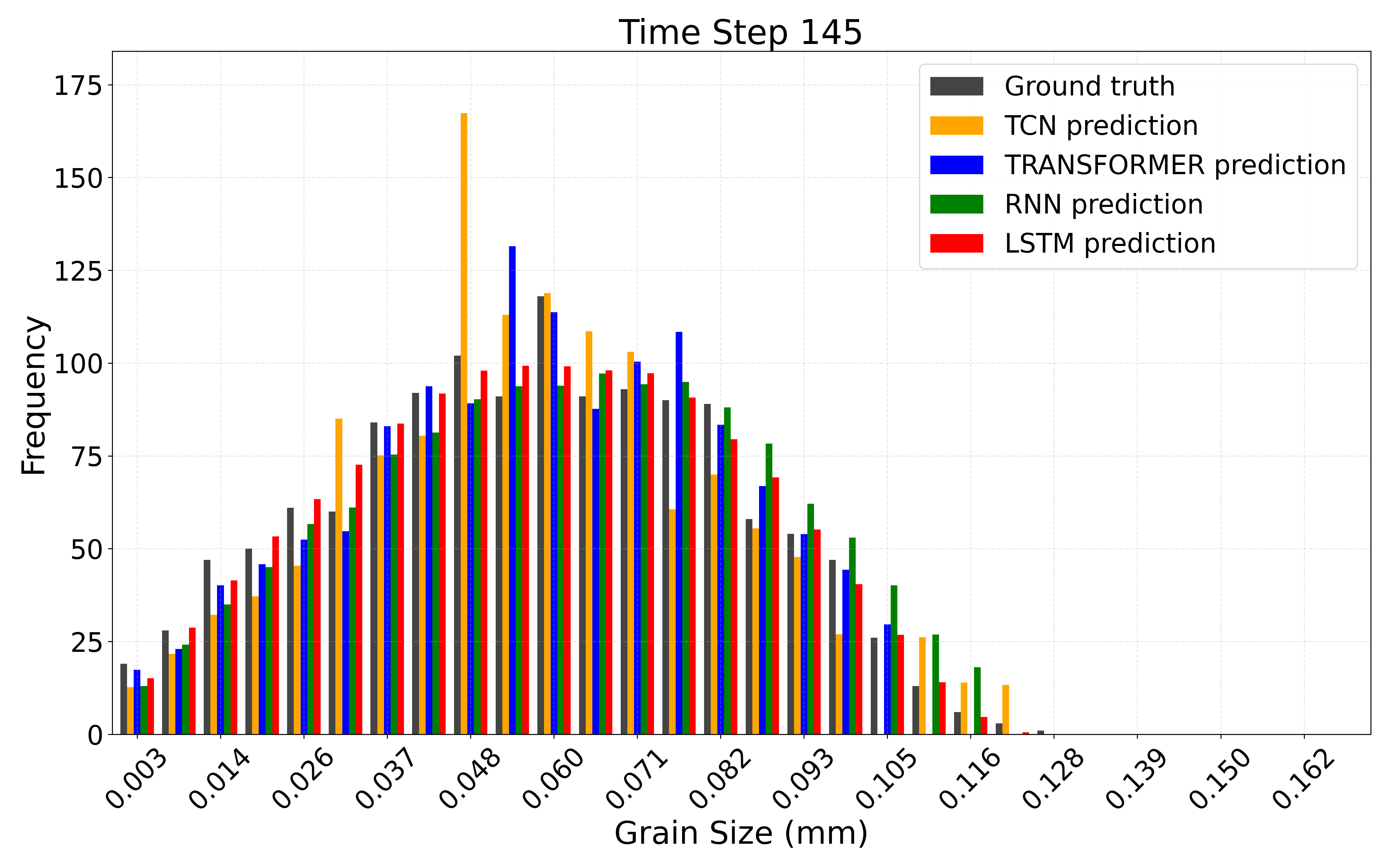}
\hfill
\includegraphics[width=0.32\textwidth]{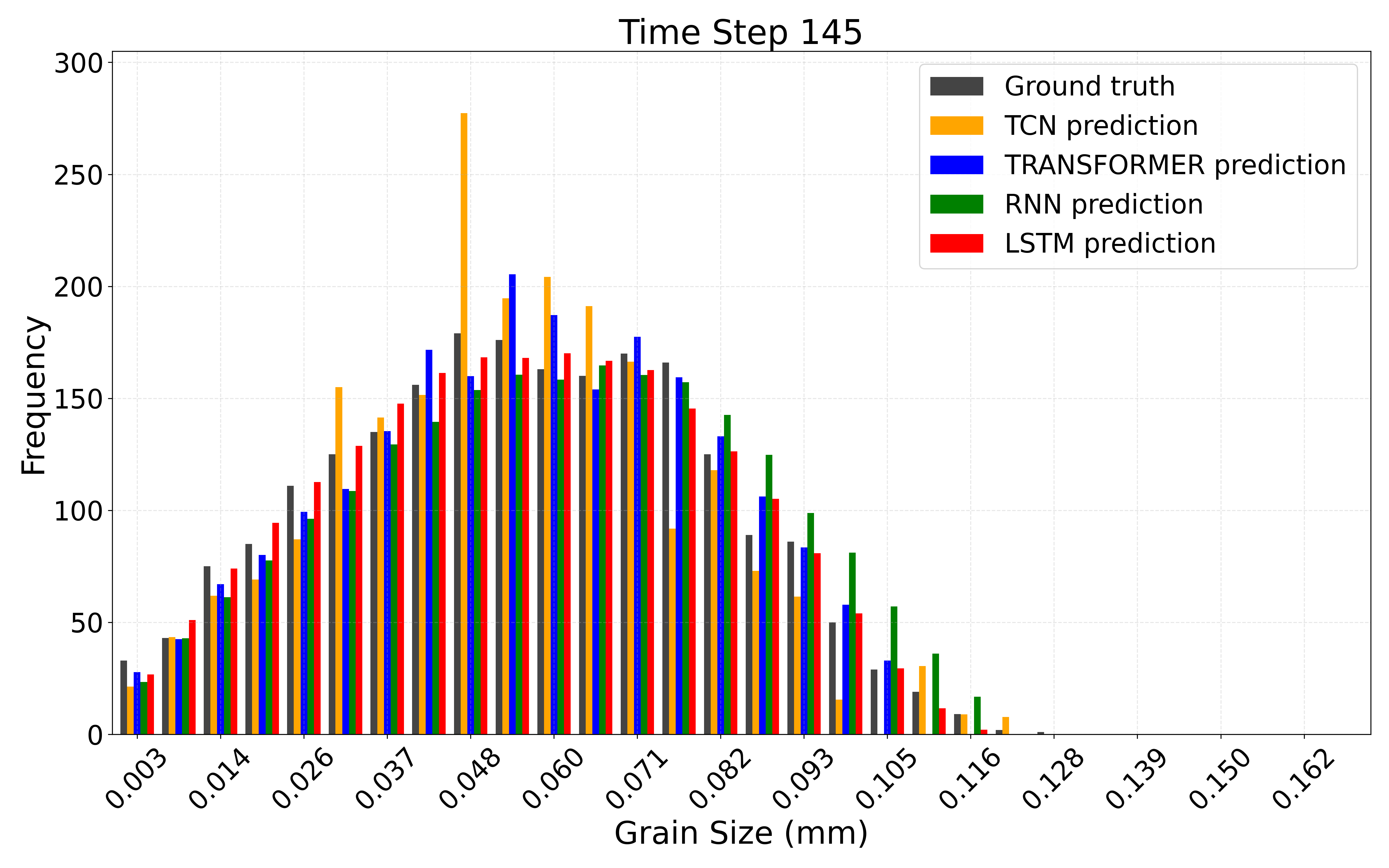}
\text{t = 145 min}\\
\vspace{0.5mm}

% ------- Row: t = 60 min -------

\includegraphics[width=0.32\textwidth]{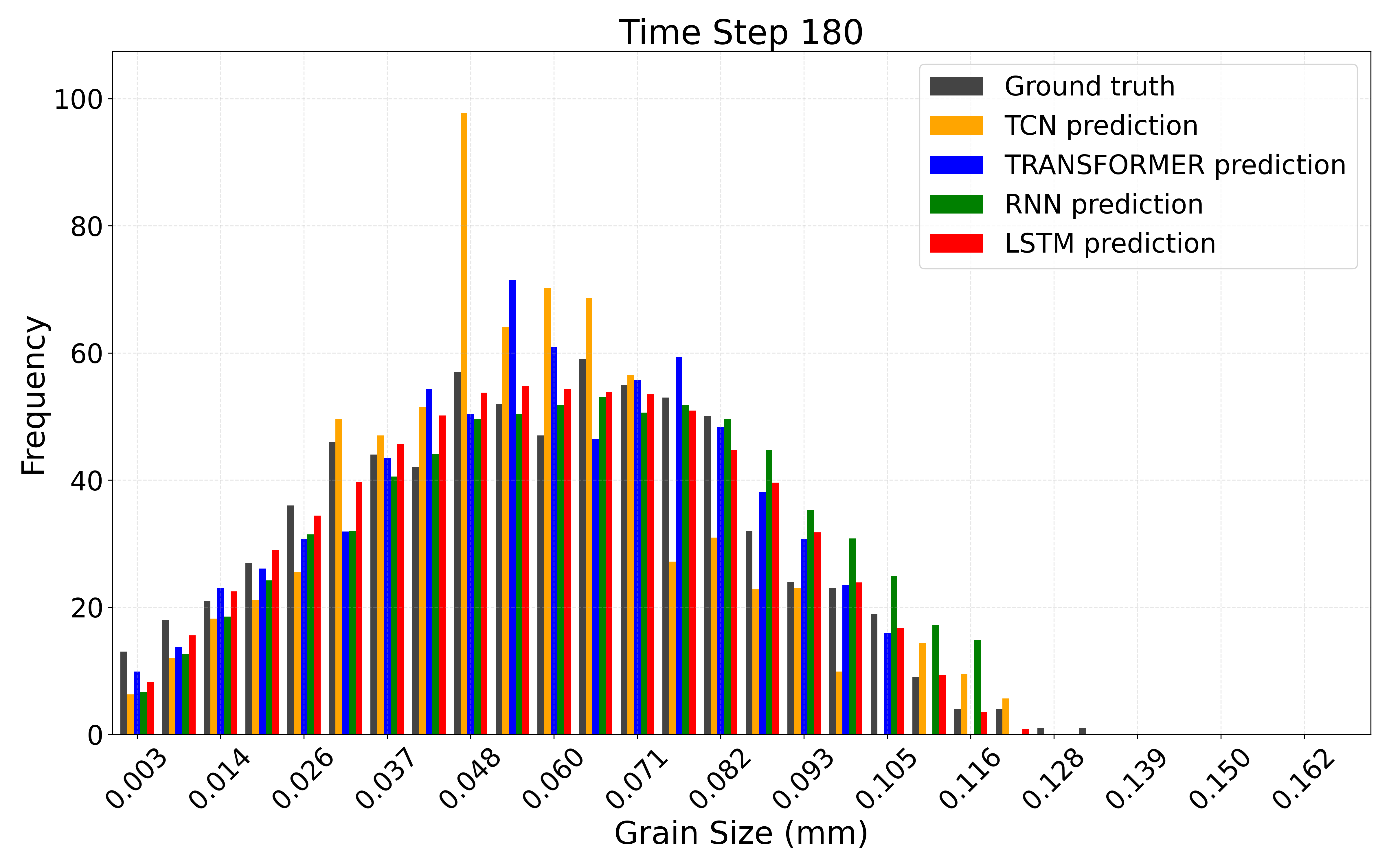}
\hfill
\includegraphics[width=0.32\textwidth]{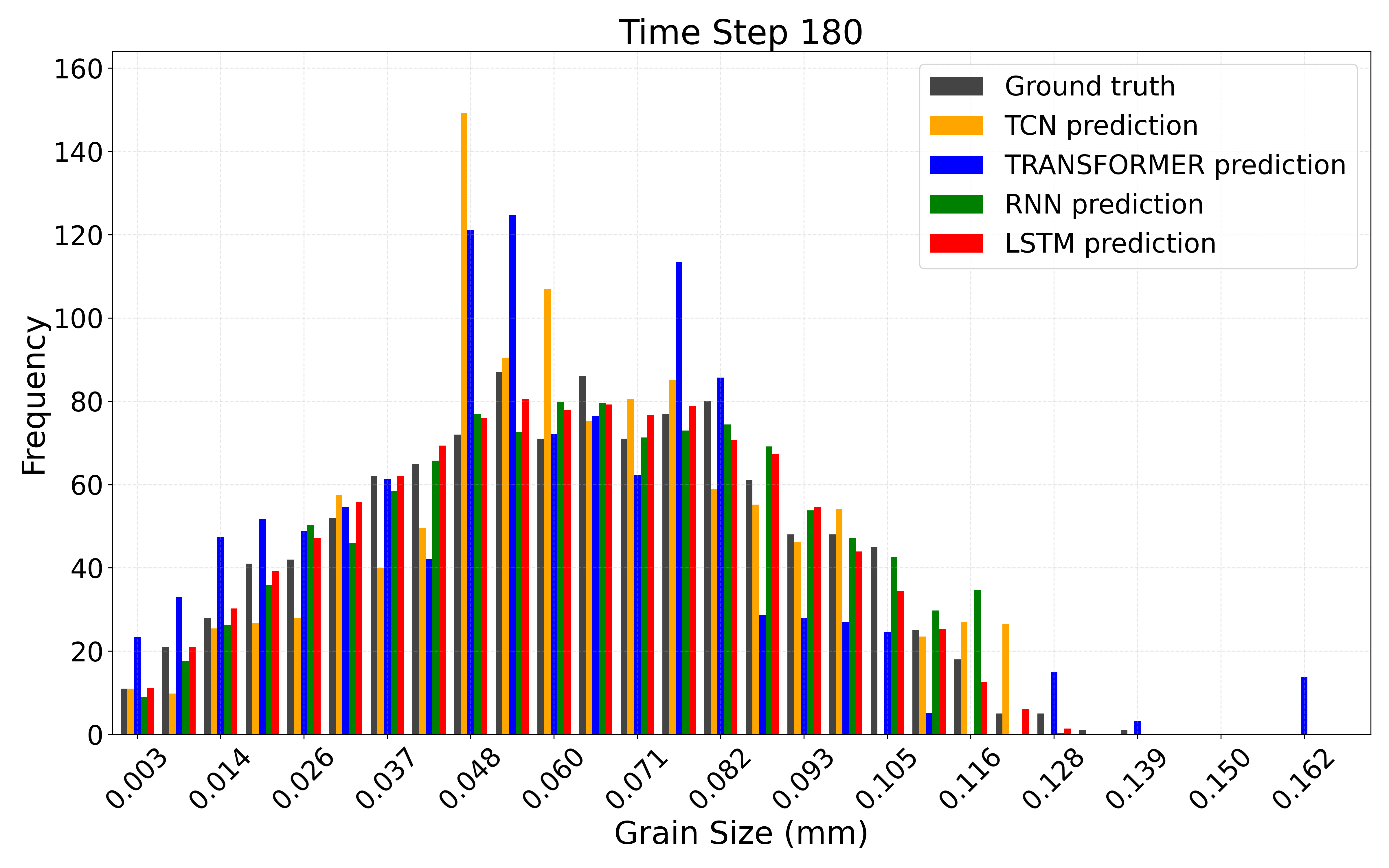}
\hfill
\includegraphics[width=0.32\textwidth]{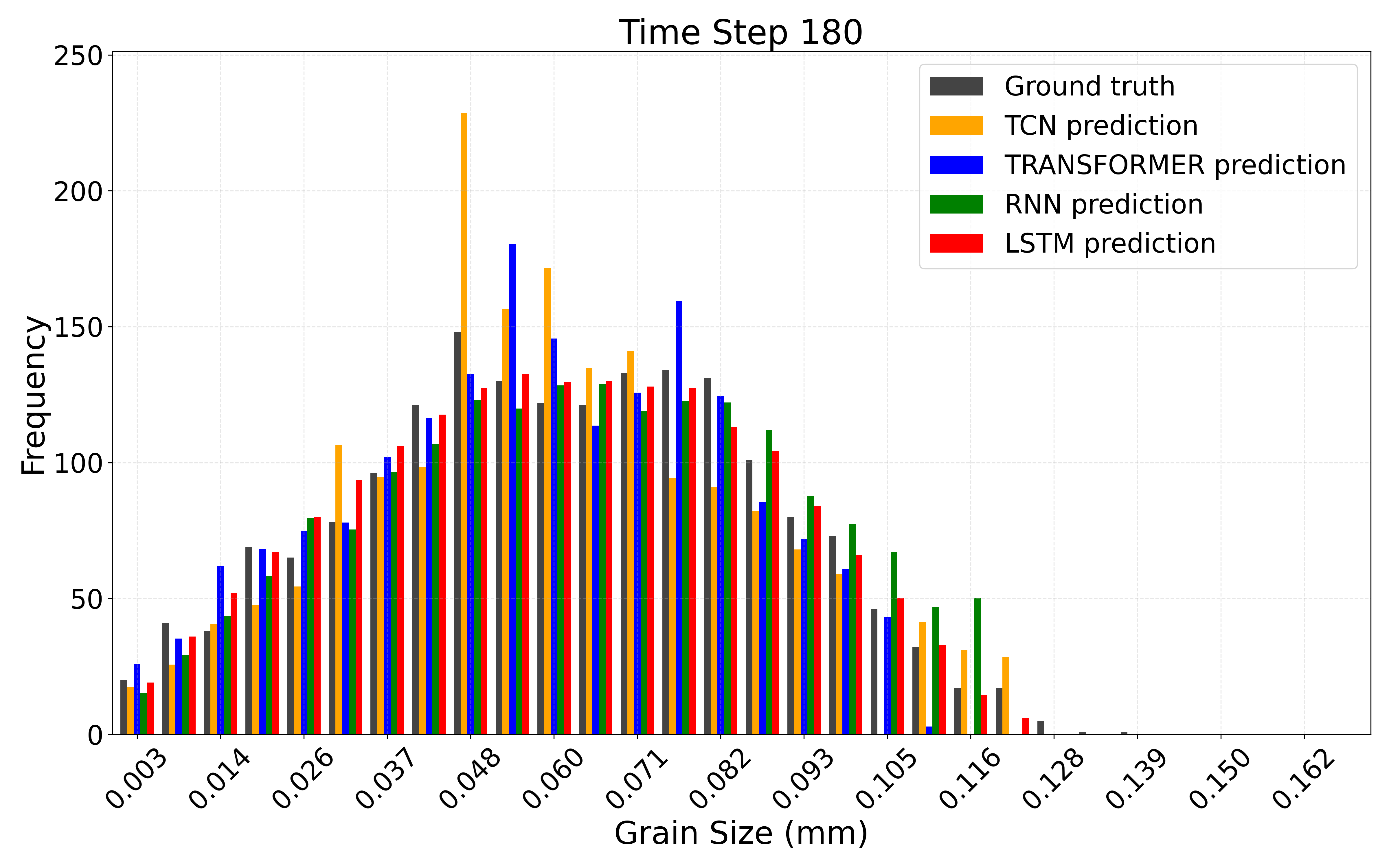}
\text{t = 180 min} \\
\vspace{0.1mm}

\caption{
Comparison of predicted and ground truth grain size distributions over time for three representative test simulations. Each column corresponds to a different spatial domain: 
(A) 3~mm~$\times$~3~mm, 
(B) 4~mm~$\times$~4~mm, and 
(C) 5~mm~$\times$~5~mm. 
Each row shows selected time steps from 6 to 180 minutes. Over the extended horizon, the LSTM maintains physically consistent predictions, whereas the Transformer, TCN, and RNN exhibit increasing divergence. All cases are taken from the unseen test set.
}
\label{fig:three_column_1}
\end{figure}

To assess the model’s generalization capability, Figure~\ref{fig:three_column_1} illustrates predictions for three representative simulation cases randomly selected from the test set with different spatial domains (3 mm × 3 mm, 4 mm × 4 mm, and 5 mm × 5 mm). These cases differ in their initial conditions, including grain count and distribution width. Following the same methodology as the one-hour analysis, this figure extends the prediction horizon to three hours (up to 180 minutes) using the same initial input window. This extended evaluation highlights the model’s ability to maintain stability and physical consistency over longer temporal scales. Despite the increased forecasting length, the LSTM model continues to reproduce realistic grain size distributions in close agreement with the TRM reference, with the mean relative error increasing moderately to about 16\%, which remains acceptable for such long-term predictions. In contrast, the Transformer and TCN architectures exhibited noticeable divergence beyond one hour, while the RNN showed intermediate behavior, maintaining reasonable accuracy for short-term forecasts but gradually losing precision over extended horizons, confirming the superior temporal stability of the LSTM. The ability to maintain consistent performance across different domain sizes, microstructural complexities, and extended prediction times further underscores the robustness and scalability of the proposed approach. These results suggest that the model captures the fundamental statistical relationships governing grain growth—such as the link between boundary curvature and migration rate—rather than memorizing specific configurations, making it suitable for broader applications including experimental datasets or simulations with varying boundary conditions.\\

From a computational perspective, the proposed DL models drastically reduce simulation time. Each 1h-TRM sequence requires approximately 20 minutes of computation on a single CPU core. In contrast, once trained, the LSTM model generates full one-hour forecasts in less than 5 seconds, representing a speed-up by a factor 200. This computational efficiency makes the approach suitable for deployment in real-time environments such as digital twins, adaptive process control, or iterative materials design frameworks. Moreover, this speed-up does not come at the expense of accuracy, as demonstrated by the consistently low error metrics across time steps. Such acceleration could significantly enhance the feasibility of large-scale parametric studies or uncertainty quantification in materials research. Integrating the trained model within multi-scale simulation pipelines could enable near-instantaneous feedback for process optimization, bridging the gap between high-fidelity physics-based simulations and data-driven surrogates.

\section{Conclusion}
This study introduced a deep learning framework for forecasting grain size distributions and modeling the microstructural evolution of polycrystalline materials during grain growth. By combining mean-field descriptors with sequence-based architectures (RNN, LSTM, TCN, and Transformer), the models were able to accurately capture the evolution of grain size over time. The proposed approach offers an efficient alternative to conventional full-field simulations, reducing computational cost while maintaining physical consistency.\\

The forecasting strategy played a central role in this performance. The recursive sliding-window method, which integrates both past observations and previously predicted values, ensured temporal coherence and improved long-term stability. Even with a short observed history, the model could continue predicting further into the future, up to three hours, while remaining stable and accurate.\\

Model accuracy was further improved by adopting a 30-bin histogram representation and an 80:15:5 data split. This configuration provided a good balance between resolution, stability, and generalization. All models performed comparably well for one-hour forecasts; however, the LSTM architecture demonstrated the most reliable long-term behavior, maintaining accuracy and stability throughout extended predictions of up to three hours, whereas the other models gradually diverged. Its capacity to preserve temporal information proved essential for accurately capturing the dynamics of grain growth.\\

A key strength of this framework is its computational efficiency and temporal scalability. While a single TRM simulation requires several minutes, the trained models generated forecasts extending up to three hours in only a few seconds. This ability to produce physically consistent predictions over extended horizons makes the approach attractive for real-time applications. Moreover, the use of compact and interpretable descriptors ensures scalability for industrial implementation.\\

Future work will extend this approach to more complex conditions, including non-isothermal and longer annealing and additional mechanisms such as GG with second-phase particles inducing Smith-Zener pinning mechanism~\cite{smith1948,zener1949} and which can be taken into account in high-fidelity TRM simulations \cite{florez2025}. The exploration of higher-order statistical descriptors also represents a promising direction. Overall, the integration of physical knowledge with data-driven forecasting opens new perspectives for efficient and intelligent modeling of microstructural evolution.

\section*{Data Availability}
The data that support the ﬁndings of this study are available from the corresponding
author upon reasonable request.

\section*{Acknowledgments}
The authors acknowledge the valuable support of ArcelorMittal, Aperam, Aubert \& Duval, CEA, Constellium, Framatome, and Safran. This work was funded in part by the French National Research Agency (ANR) through the DIGIMU consortium and the RealIMotion Industrial Chair (Grant No. ANR-22-CHIN-0003).

 % \clearpage
\printbibliography

\end{document}